\documentclass{article} 
\usepackage{iclr2027_conference,times}

\usepackage{amsmath,amsfonts,bm}

\def\eqref#1{equation~\ref{#1}}

\def\1{\bm{1}}

\DeclareMathAlphabet{\mathsfit}{\encodingdefault}{\sfdefault}{m}{sl}
\SetMathAlphabet{\mathsfit}{bold}{\encodingdefault}{\sfdefault}{bx}{n}

\usepackage[table]{xcolor}
\usepackage{booktabs}
\usepackage{colortbl}
\usepackage{diagbox}
\usepackage{hyperref}
\usepackage{url}
\usepackage{xspace}
\usepackage{enumitem}
\usepackage{graphicx}
\usepackage{amsthm}
\usepackage{amssymb}
\usepackage{comment}

\usepackage{booktabs}
\usepackage{multirow}
\usepackage{xcolor}
\usepackage{graphicx}
\usepackage{pifont}

\usepackage{wrapfig}
\usepackage{subcaption}
\usepackage{algorithm}
\usepackage{algorithmic}
\usepackage{tcolorbox}
\tcbuselibrary{skins}

\newtheoremstyle{mydefinition}
  {6pt}                    
  {6pt}                    
  {\normalfont}            
  {}                       
  {\bfseries\scshape}      
  {.}                      
  {0.5em}                  
  {}                       

\theoremstyle{mydefinition}
\newtheorem{definition}{Definition}
\newcommand{\method}{\textsc{WaG}\xspace}

\title{World-as-Graph: Relational World Modeling Through Latent Space Graphs}

\author{
Yaqi Yang$^{1}$,
Shuo Huang$^{2,4}$,
Yujin Huang$^{3}$,
Fucai Ke$^{2}$,
Jiatong Han$^{4}$,
Xin Zheng$^{1}$\thanks{Corresponding author.}  \\
$^{1}$School of Computing Technologies, RMIT University, Australia \\
$^{2}$Faculty of Information Technology, Monash University, Australia \\
$^{3}$School of Computing and Information Systems, The University of Melbourne, Australia \\
$^{4}$KStelrix Star Dynamics Lab, KStelrix, China \\
\texttt{s4219362@student.rmit.edu.au} \quad
\texttt{xin.zheng2@rmit.edu.au}
}

\iclrfinalcopy 
\begin{document}

\maketitle

\begin{abstract}
World models aim to learn representations of real-world environments and predict their future evolution. Recent object-centric world models have made expressive progress by representing visual scenes as sets of object-level latent states, but object-object relations are often captured only implicitly, which limits 
\textit{explicit relational and temporal structure modeling} and \textit{object-centric dynamic memory modeling}. To address such challenges, we propose \textbf{\underline{\textsc{W}}}orld-\textbf{\underline{\textsc{a}}}s-\textbf{\underline{\textsc{G}}}raph (\textbf{\method}), a graph-based object-centric world model that introduces relational inductive bias into JEPA-style predictive representation learning. The proposed \method contains two main modules: (1) Relation-aware structure induction, which constructs time-varying latent graphs from object-centric slots and designs relation-aware object masking policies to guide relational object representation learning in latent space; (2) Object-centric memory transition, which maintains and updates object-level dynamic states by combining relational information from neighboring objects with historical memory, enabling effective autoregressive future prediction. Extensive experiments on both visual reasoning and robotic manipulation tasks demonstrate the superior performance of our proposed \method. Our code is available at https://github.com/Scarlett-Yyq/World-as-Graph.
\end{abstract}

\section{Introduction}
World models~\citep{ha2018world,ding2025understanding,lecun2022path} have recently attracted increasing attention for capturing and modeling real-world environments and predicting their future evolutions. By learning future states from past observations, world models provide better world understanding, prediction, and decision-making across a wide range of applications, from video generation to robotics~\citep{cho2024sora,bruce2024genie,wu2024ivideogpt,wang2026unified,yang2026chain,zhang2026dreamvla,bi2026motus}.
In this context, a key aspect of world modeling is to learn predictive and structural representations that capture world dynamics in a \textbf{latent space} by explicitly modeling how individual objects interact and evolve over time~\citep{yang2026chain,lei2026spartan,feng2026learning,kipf2020contrastive}.

Inspired by this perspective, the joint-embedding predictive architecture, i.e., JEPA~\citep{lecun2022path,assran2023self,klindt2026does,bai2026temporal}, has demonstrated impressive progress for predictive representation learning in world models~\citep{saito2025point,yang2026chain}. Instead of reconstructing observations at the pixel level, JEPA predicts future representations directly in latent space, supporting effective representation learning for perception and action.
Meanwhile, to better capture object interactions in the world, object-centric models have recently been adopted as a foundation for world modeling~\citep{nam2026causal}. These models typically represent a visual scene as a set of object-level latent states (i.e., slots), allowing different objects to be modeled separately while capturing visual dynamics and learning object-centric predictive representations~\citep{mosbach2025sold,nishimoto2026object,kipf2022conditional,ferraro2025focus}.

Despite the promising performance of object-centric representation learning with JEPA, recent studies suggest that object-centric predictors may not guarantee that meaningful interactions are explicitly captured from object slots alone~\citep{lei2026spartan}. Existing object-centric world models, including Causal-JEPA~\citep{nam2026causal}, therefore introduce different mechanisms to encourage interaction learning, such as structured dynamics, sparse attention, and task-specific supervision~\citep{kipf2020contrastive,feng2026learning,mosbach2025sold,ferraro2025focus}. However, without explicit relational inductive biases, models may still predict future states primarily from object self-dynamics, leaving two essential challenges: 
\textbf{C1: Limited relational and temporal structure modeling}, where existing object-centric latent world models operate on latent object slots, but often leave object-object relations implicit through attention or object-level masking. In particular, they do not distinguish objects according to relational roles or explicitly model how object relations evolve over time.
As a result, masked latent prediction in JEPA may provide limited self-supervised signals for effectively learning object interactions.
\textbf{C2: Limited object-centric dynamic memory modeling}, where future prediction requires each object to carry its own interaction history together with relational information accumulated from other objects. Without explicit object-level state memory modeling, the model may lose important historical cues about how objects move, interact, and influence future states, leading to suboptimal long-horizon prediction.
\begin{wrapfigure}{r}{0.65\columnwidth}
    \centering
    \includegraphics[width=\linewidth]{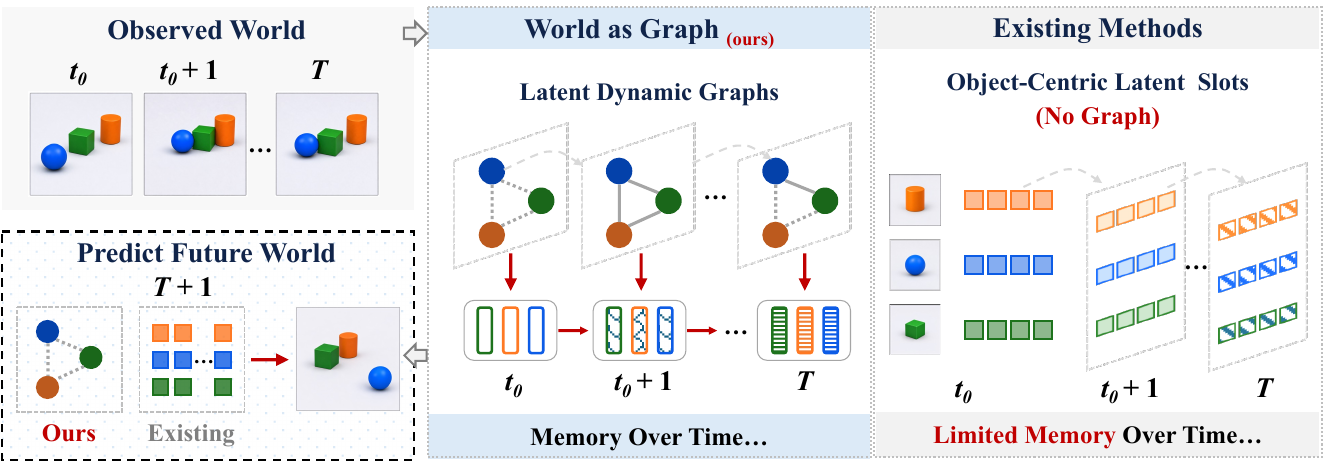}
    \vspace{-15pt}
    \caption{Comparison of our proposed World-as-Graph (\method) with existing object-centric world models \scriptsize{(with no explicit graph structures (C1) and limited temporal memory (C2)).}}
    \label{fig:intro}
    \vspace{-15pt}
\end{wrapfigure}

To address these challenges, we propose a \textbf{\underline{\textsc{W}}}orld-\textbf{\underline{\textsc{a}}}s-\textbf{\underline{\textsc{G}}}raph model, dubbed \textbf{\method}, for relational world modeling through graph-based representation learning in the latent space, as shown in Figure~\ref{fig:intro}.
The key idea is to exploit \textbf{graph structure} as a relational inductive bias (RIB)~\citep{battaglia2018relational} for object-centric JEPA-style world models, enabling structured state representations of objects and their interactions. 
Specifically, our proposed \method contains two core sub-modules: (1) \textit{Relation-aware structure induction}, which induces relational structures in latent space through a \underline{(1-a)} latent dynamic graph constructor and a \underline{(1-b)} relation-aware object masker; and (2) \textit{Object-centric memory transition}, which models the temporal evolution of object-level dynamic states through a \underline{(2-a)} temporal graph neural network (GNN) transition encoder and a \underline{(2-b)} memory-driven future predictor. 
Concretely, the latent dynamic graph constructor treats object slots as nodes and builds time-varying relational graphs, followed by the relation-aware object masker, which selects objects based on relational centrality or temporal neighborhood changes. Masked object states are recovered from relational context and historical information. The first module models how object relations and structural dependencies evolve over time and guides JEPA-style predictive learning, addressing challenge \textbf{C1}. Afterward, the temporal GNN transition encoder updates each object state using neighboring relations and its historical memory. The memory-driven future predictor recursively evolves these memory states and constructs future dynamic graphs to predict future object states. The second module models how object-centric dynamic states evolve and supports future world prediction, addressing challenge \textbf{C2}.
Extensive experiments on visual reasoning and robotic manipulation tasks show that our proposed \method improves performance by up to 30.03 percentage points, while achieving $2.4\times$ faster training and up to $16.4\times$ lower peak GPU memory.
In summary, the contributions of this work are listed below:
\begin{itemize}[leftmargin=1em, topsep=0pt, parsep=0pt, partopsep=0pt]
    \item To the best of our knowledge, we are the first to introduce latent space graphs for relational modeling in object-centric JEPA-style world models to explicitly capture object interactions and temporal dynamics in a \textbf{\underline{\textsc{W}}}orld-\textbf{\underline{\textsc{a}}}s-\textbf{\underline{\textsc{G}}}raph (\textbf{\method}).
    \item The proposed \method consists of two core sub-modules that jointly conduct (1) \textit{relation-aware structure induction} composed of a latent dynamic graph constructor and a relation-aware object masker, and (2) \textit{object-centric memory transition} composed of a temporal GNN transition encoder and a memory-driven future predictor, for expressive future world prediction.
    \item Extensive experiments demonstrate the effectiveness and benefits of \method in introducing relational inductive bias to model object interactions and future dynamics.
\end{itemize}

\textbf{Prior Work.} Our work is closely related to three lines of research: (a) \textit{joint-embedding predictive architectures (JEPA)}, (b) \textit{graph world models}, and, to a lesser extent, (c) \textit{dynamic graph representation learning}. Specifically, \textit{JEPA methods}~{\footnotesize{\citep{lecun2022path,assran2023self,klindt2026does,bai2026temporal,
saito2025point,tuncay2025audio,balestriero2025lejepa,yang2026chain,nam2026causal}}} learn predictive representations directly in latent space, while recent object-centric JEPA-style world models~{\footnotesize{\citep{feng2026learning,wang2025dyn}}} further represent visual scenes as sets of object-level latent states for learning object-wise dynamics. However, these existing methods often capture implicit interactions, while our \method explicitly constructs evolving latent dynamic graphs for object state interactions. 
Most existing \textit{graph world models}~{\footnotesize{\citep{liu2026graph,feng2025graph,song2026understanding,
wang2026structural}}} introduce world-modeling principles into graph-structured data, i.e., \textit{world modeling for graphs}, while our work first introduces latent graph structures as relational inductive biases for object-centric world modeling, i.e., \textit{graphs for world modeling}.
\textit{Dynamic graph representation learning}~\citep{feng2025comprehensive, zheng2025survey} is closely related to the temporal GNN component in \method.
More detailed discussions of these related research directions are provided in Appendix~\ref{appx:related_work}.
\section{Methodology}
\subsection{Preliminary}
\paragraph{Notation.}
Given a video representing a dynamic world as an image sequence,
a pretrained object-centric encoder $f_{\Phi^{*}}$ extracts $N$ latent slot representations from each frame at time step $t$ as $\mathbf{S}_t = [\mathbf{s}_t^1,\ldots,\mathbf{s}_t^N]^{\top} \in \mathbb{R}^{N \times D_{0}}$, where $\mathbf{s}_t^i \in \mathbb{R}^{D_{0}}$ denotes the latent feature of the $i$-th slot, $D_{0}$ is the feature dimension, and $N$ is the fixed number of object slots. 
In this work, we transfer the latent slot representations to a discrete-time dynamic graph sequence $\mathcal{G}=\{G_t\}_{t\in\mathcal{T}}$. Each graph snapshot at time step $t$ is defined as $G_t=(\mathcal{V}_t,\mathcal{E}_t,\mathbf{S}_t)$, where $\mathcal{V}_t=\{v_t^1,\ldots,v_t^N\}$ denotes the set of object nodes in the latent space, and $\mathcal{E}_t \subseteq \mathcal{V}_t \times \mathcal{V}_t$ denotes the set of edges. Each node $v_t^i$ corresponds to the $i$-th object-centric slot at time step $t$ and is associated with the latent feature $\mathbf{s}_t^i$. 
For future prediction, we have $\mathcal{T}=\mathcal{T}_{\mathrm{hist}}\cup\mathcal{T}_{\mathrm{pred}}$, where $T_h$ and $T_p$ indicate the history window length and future prediction horizon, respectively. Given the current time step $t$, the corresponding history and prediction time sets are defined as $\mathcal{T}_{\mathrm{hist}}=\{t-T_h+1,\ldots,t\}$ and $\mathcal{T}_{\mathrm{pred}}=\{t+1,\ldots,t+T_p\}$.
Given the observed slot states $\{\mathbf{S}_{t}\}_{t\in\mathcal{T}_{\mathrm{hist}}}$, the world model aims to predict the future slot states $\{\mathbf{S}_{t}\}_{t\in\mathcal{T}_{\mathrm{pred}}}$.
\paragraph{Problem Setting.} 
Following the world model framework of \citet{ha2018world}, a model learns a compact representation of the environment from observations and predicts how that representation evolves in response to current actions. In this work, we follow this general formulation together with the object-centric and JEPA-style representation learning paradigm~\citep{nam2026causal, assran2023self}, and introduce graph machine learning and relational inductive bias to explicitly model interactions within the latent world states. Based on this, we further define a graph-structured world model, termed the \textbf{\underline{\textsc{W}}}orld-\textbf{\underline{\textsc{a}}}s-\textbf{\underline{\textsc{G}}}raph (\textbf{\method}) model, as follows.
\begin{definition}[\textsc{\textbf{World-as-Graph Model}}]
Given the historical object-centric latent states
$\{\mathbf{S}_t\}_{t\in\mathcal{T}_{\mathrm{hist}}}$,
a World-As-Graph (\method) model is defined as
$\mathcal{M}_{\mathrm{WAG}}=\langle \mathcal{R},\mathcal{F}_{\theta,\omega}\rangle$, where:
\begin{itemize}[leftmargin=1em, topsep=0pt, parsep=0pt, partopsep=0pt]
    \item \textit{Relation-Aware Structure Induction $\mathcal{R}=\langle\Gamma_{\psi},\pi_{\mathrm{mask}}\rangle$}: first constructs dynamic relational graphs from object-centric latent representations through the latent dynamic graph constructor $\Gamma_{\psi}$, and then uses the relation-aware masking policy $\pi_{\mathrm{mask}}$ to identify relationally informative object nodes based on the graph structure.
    \item \textit{Object-Centric Memory Transition $\mathcal{F}_{\theta,\omega}=\langle f_{\mathrm{enc}}^{\theta},f_{\mathrm{pred}}^{\omega}\rangle$}: 
    maintains and updates object-level memory states from the masked historical graph sequence through the temporal GNN transition encoder $f_{\mathrm{enc}}^{\theta}$, and further evolves these memory states through the memory-driven future predictor $f_{\mathrm{pred}}^{\omega}$ to predict the future object-centric latent states $\{\widehat{\mathbf{S}}_t\}_{t\in\mathcal{T}_{\mathrm{pred}}}$.
\end{itemize}
\end{definition}
\begin{figure}[!t]
\begin{center}
\vspace{-5pt}
\includegraphics[width=\linewidth]{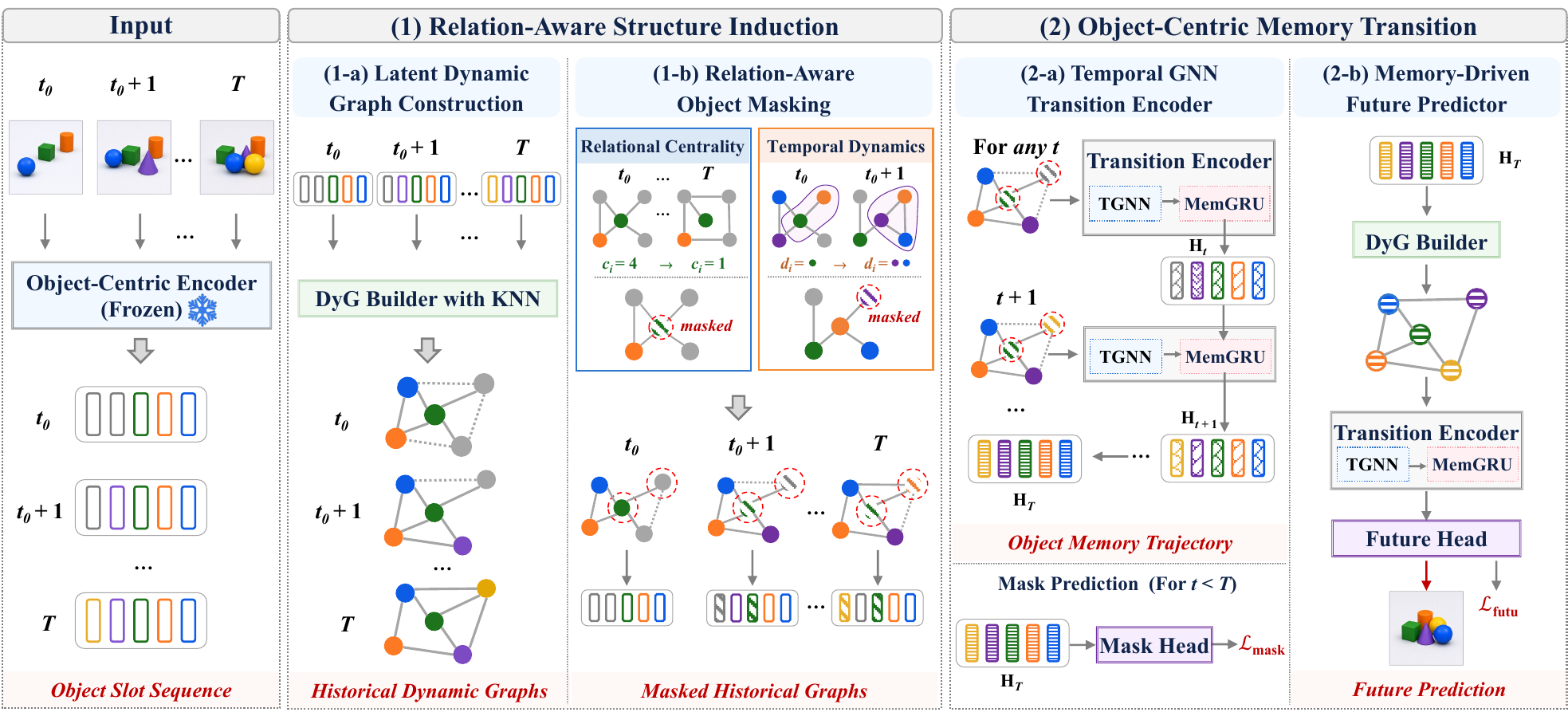}
\end{center}
\vspace{-10pt}
\caption{Overall framework of our proposed \method.}
\vspace{-15pt}
\label{fig:framework_gow}
\end{figure}
\subsection{World-as-Graph Framework}
To explicitly model structured state representations of objects and their interactions, in this work, we first propose \method, a graph-based object-centric world model.
The overall framework is presented in Fig.~\ref{fig:framework_gow}.
Given an input video, a frozen object-centric encoder first extracts object-level latent representations over time. The proposed \method then performs two main stages. First, \textit{relation-aware structure induction} constructs latent dynamic graphs among object slots and applies relation-aware masking based on relational centrality or temporal dynamics. Second, \textit{object-centric memory transition} propagates relational information through a temporal GNN and updates object-level memory states over the observed history. These memory states are used for masked-state prediction during training and are further evolved by the memory-driven future predictor to autoregressively generate future object representations in latent space.
\subsubsection{Relation-Aware Structure Induction}
This module explicitly induces relational bias through dynamic graph structure learning among object-centric slots and further guarantees relation-aware representation learning. Specifically, it contains two core sub-components: (a) \textit{Latent dynamic graph construction}, which treats each object slot as a node and builds temporal object-object edges based on latent representation similarity; (b) \textit{Relation-aware object masking}, which leverages the relational structure to inform the model of which objects deserve more attention.
\paragraph{Latent Dynamic Graph Construction.}
To explicitly model relational structures in latent space, given the object slot representations $\mathbf{S}_t = [\mathbf{s}_t^1,\ldots,\mathbf{s}_t^N]^{\top}$, each slot $\mathbf{s}_t^i$ is treated as an object node $v_t^i$. Our proposed \method constructs a latent dynamic graph through a parameter-free dynamic graph builder $\Gamma$ based on a similarity function $\mathrm{sim}(\cdot,\cdot)$ and K-nearest neighbor (KNN) algorithm as:  
\begin{equation}\label{eq:dynamic_graph}
G_t=\Gamma_{K}^{{\mathrm{sim}}}\left(\mathbf{S}_t\right)=\left(\mathcal{V}_t,\mathcal{E}_t,\mathbf{S}_t\right),\quad \mathcal{E}_t=\left\{(i,j)\mid j\in\operatorname*{arg\,Top-\mathit{K}}_{k\neq i}{\mathrm{sim}}\left(\mathbf{s}_i^t,\mathbf{s}_k^t\right)\right\},
\end{equation}
where $\mathcal{E}_t$ denotes the set of edges connecting each object to its Top-$K$ most similar nodes. In this work, we use cosine similarity, and the edge weight is given by $w_{ij}^t=\operatorname{cos}\left(\mathbf{s}_i^t,\mathbf{s}_j^t\right)$. This produces a sequence of time-varying latent graphs as $\mathcal{G}=\{G_t\}_{t\in\mathcal{T}}$.
\paragraph{Relation-Aware Object Masking.}
For each latent dynamic graph $G_t$, object nodes are selected for masking according to their relational roles in the constructed graph. In this way, the proposed \method is able to recover the masked object states by jointly using relational information from other objects and the object's own historical states. Specifically, let $\mathcal{I}_{\mathrm{mask}}\subset\{1,\ldots,N\}$ denote the selected masked node indices, with $|\mathcal{I}_{\mathrm{mask}}|=M$ and $M < N$. Unlike random slot masking, our relation-aware object masking strategy is guided by the latent dynamic graph topology over the history window. Specifically, we design two strategies as below. 

\textit{(I) Relational Centrality Masking.} 
We first select relationally central object nodes according to their roles in the constructed latent dynamic graphs. Since each node in the KNN graph has a fixed out-degree $K$, we use the in-degree as the centrality signal to measure how frequently an object node $v_t^{i}$ appears in the neighbor sets $\mathcal{N}_t^{j}$ of other objects over the observed history $\mathcal{T}_{\mathrm{hist}}$.
Therefore, we define the relational centrality score $c_i$ and select the masked node indices $\mathcal{I}_{\mathrm{mask}}^{\mathrm{cen}}$ as
\begin{equation} \label{eq:degree_mask}
c_i = \sum_{t\in\mathcal{T}_{\mathrm{hist}}} \sum_{\substack{j=1\\ j\neq i}}^{N} \mathbb{I}\!\left[i\in\mathcal{N}_t^{j}\right], \qquad \mathcal{I}_{\mathrm{mask}}^{_\mathrm{cen}} = \operatorname*{arg\,Top-\mathit{M}}_{i\in\{1,\ldots,N\}} c_i , \end{equation}
where the Top-$M$ objects with the highest centrality scores are selected for masking. 
By recovering these relationally central object states in JEPA-style learning, the proposed \method exploits relational information from other objects together, rather than relying mainly on object self-dynamics.

\textit{(II) Temporal Dynamics Masking.}
We select object nodes according to the temporal changes of their relational contexts. The intuition behind is straightforward: if an object's neighborhood changes frequently across consecutive time steps, its state is more likely to be affected by evolving interactions with other objects. Therefore, we define the temporal dynamics score $d_i$ by measuring the accumulated changes in the neighbor set $\mathcal{N}_t^i$ over the observed history $\mathcal{T}_{\mathrm{hist}}$, and select the masked node indices $\mathcal{I}_{\mathrm{mask}}^{\mathrm{dyn}}$ as
\begin{equation}\label{eq:temporal_mask}
d_i=\sum_{t\in\mathcal{T}_{\mathrm{hist}}}\left(K-\left|\mathcal{N}_{t}^{i}\cap\mathcal{N}_{t-1}^{i}\right|\right),\qquad \mathcal{I}_{\mathrm{mask}}^{_\mathrm{dyn}}=\operatorname*{arg\,Top-\mathit{M}}_{i\in\{1,\ldots,N\}} d_i,
\end{equation}
where $K-\left|\mathcal{N}_{t}^{i}\cap\mathcal{N}_{t-1}^{i}\right|$ measures the number of changed neighbors between two consecutive time steps. By recovering these temporally dynamic object states, the proposed \method exploits evolving relational information across time.

After obtaining $\mathcal{I}_{\mathrm{mask}} \in \left\{\mathcal{I}_{\mathrm{mask}}^{_\mathrm{cen}}, \mathcal{I}_{\mathrm{mask}}^{_\mathrm{dyn}}\right\}$, we replace the corresponding object features with a learnable mask token over the observed history, while keeping the first historical graph $G_{t_0}$ at $t_0=t-T_h+1$ fully visible to initialize the object states. This produces a partially observed historical graph sequence $\mathcal{G}^{\dagger}=\{G_t^{\dagger}\}_{t\in\mathcal{T}_{\mathrm{hist}}\setminus\{t_0\}}$, where we have $G_t^{\dagger}=(\mathcal{V}_t,\mathcal{E}_t,\mathbf{S}_t^{\dagger})$ with masked object representations $\mathbf{S}_t^{\dagger}$. One goal of the proposed \method is to recover the masked object states from their relational context and historical information. 


\subsubsection{Object-Centric Memory Transition}
Although the relation-aware structure induction explicitly models object-object relations \textit{within} each time step, future world prediction further requires each object to maintain its own historical state across time. This becomes critical in two cases. First, a masked object node cannot observe its own state in the remaining historical frames. Second, during future generation through autoregressive rollout, no future observations are available for any object node. 
In light of this, we design an object-centric memory transition module, where each object node maintains an individual \textit{memory state} to preserve its historical information and relational interactions over time. Specifically, this module contains two sub-components: (a) \textit{Temporal GNN transition encoder}, which updates object-level memory states over the observed history through graph structure-guided message passing, and (b) \textit{Memory-driven future predictor}, which evolves the learned memory states to predict future object states without new observations.
\paragraph{Temporal GNN Transition Encoder.}
Given the partially observed historical graph $G_t^{\dagger}$ after relation-aware object masking, we maintain each object node $v_t^i$ an individual memory state $\mathbf{h}_t^i\in\mathbb{R}^{D_1}$, where $D_1$ is the memory feature dimension. Therefore, we define the object-level memory states as $\mathbf{H}_t=[\mathbf{h}_t^1,\ldots,\mathbf{h}_t^N]^{\top} \in \mathbb{R}^{N \times D_{1}}$. Since the first historical graph at $t_0$ is kept fully visible, the object memory states are initialized as the observed object representations $\mathbf{H}_{t_0}=\mathbf{S}_{t_0}$.

For the remaining historical time steps, we design a temporal GNN transition encoder $f_{\mathrm{enc}}^{\theta}$ that updates the object-level memory states by jointly considering the current partially observed graph $G_t^{\dagger}$ and the previously accumulated memory states $\mathbf{H}_{t-1}$ through temporal graph message passing:
\begin{equation}\label{eq:f_enc}
\mathbf{H}_t
=
f_{\mathrm{enc}}^{\theta}
\left(
G_t^{\dagger},
\mathbf{H}_{t-1}
\right)
=
\operatorname{MemGRU}_{\theta_m}
\left(
\operatorname{TGNN}_{\theta_g}\left(G_t^{\dagger}\right),
\mathbf{H}_{t-1}
\right),
\qquad
t\in\mathcal{T}_{\mathrm{hist}}\setminus\{t_0\},
\end{equation}
where $\theta=\{\theta_g,\theta_m\}$ denotes the parameters of the temporal GNN and memory-based gated recurrent unit (GRU). In this way, each object memory state $\mathbf{h}_t^i$ integrates the current object information, relational context from neighboring objects, and its previous memory state $\mathbf{h}_{t-1}^i$, therefore preserving both object-specific historical information and relational information accumulated over time.

To further encourage our proposed model to learn predictive object representations through the masking policy, we apply an auxiliary masked-state prediction head $g_{\mathrm{mask}}^{\epsilon}$ to the corresponding memory states for generating predicted masked object representations as:
\begin{equation}\label{eq:mask_pred_representation}
\widehat{\mathbf{S}}_{t}^{_\mathrm{mask}}
=
g_{\mathrm{mask}}^{\epsilon}
\left(
\mathbf{H}_{t},\mathcal{I}_{\mathrm{mask}}
\right),
\qquad
t\in\mathcal{T}_{\mathrm{hist}}\setminus\{t_0\}.
\end{equation}
\paragraph{Memory-Driven Future Predictor.}
After the observed history ends, we design a memory-driven future predictor $f_{\mathrm{pred}}^{\omega}$ to autoregressively evolve the learned object-level memory states using the previously learned memory for relational future prediction. Specifically, for each future time step $t\in\mathcal{T}_{\mathrm{pred}}$, we first apply the same latent dynamic graph constructor to the previous memory states and the same temporal GNN encoder to capture the object-level interactions. Then, the designed memory-driven future predictor $f_{\mathrm{pred}}^{\omega}$ evolves the previous memory states with temporal information to update the future memory states, so we could generate the future object representation with transitioned memory as:
\begin{equation}\label{eq:f_pred}
\widehat{G}_{t-1}=\Gamma_{K}^{\mathrm{sim}}\left(\mathbf{H}_{t-1}\right),
\qquad
\mathbf{H}_{t}
=
f_{\mathrm{pred}}^{\omega}
\left(
\widehat{G}_{t-1},
\mathbf{H}_{t-1}
\right),
\qquad
\widehat{\mathbf{S}}_{t}^{_\mathrm{futu}}
=g_{\mathrm{proj}}^{\eta}
\left(
\mathbf{H}_{t}
\right),
\qquad
t\in\mathcal{T}_{\mathrm{pred}},
\end{equation}
where $g_{\mathrm{proj}}^{\eta}$ is a projection head for generating future object states from the updated memory states.
In this way, the proposed \method progressively constructs future relational structures and evolves object-level memory, predicting the future object-centric latent states $\{\widehat{\mathbf{S}}_t^{_\mathrm{futu}}\}_{t\in\mathcal{T}_{\mathrm{pred}}}$.
\subsubsection{Learning Objective}
\paragraph{Training Stage.}
The proposed \method is trained with two predictive objectives: (1) Historical masked-state prediction loss $\mathcal{L}_{\mathrm{mask}}$, to encourage \method to recover the original latent representations of the masked objects from their learned memory states; (2) Future-state prediction loss $\mathcal{L}_{\mathrm{futu}}$, to predict the future object-centric latent states through autoregressive memory transition with latent dynamic graph guidance. Given the predicted masked representations $\widehat{\mathbf{S}}_{t}^{_\mathrm{mask}}$ and the predicted future states $\widehat{\mathbf{S}}_{t}^{_\mathrm{futu}}$, we define the following training objectives:
\begin{equation}\label{eq:lmask_lfutu}
\mathcal{L}_{\mathrm{mask}}
=
\frac{1}{|\mathcal{T}_{\mathrm{hist}}\setminus\{t_0\}|}
\sum_{t\in\mathcal{T}_{\mathrm{hist}}\setminus\{t_0\}}
\ell
\left(
\widehat{\mathbf{S}}_{t}^{_\mathrm{mask}},
\mathbf{S}_{t}^{_\mathrm{mask}}
\right), 
\quad
\mathcal{L}_{\mathrm{futu}}
=
\frac{1}{|\mathcal{T}_{\mathrm{pred}}|}
\sum_{t\in\mathcal{T}_{\mathrm{pred}}}
\ell
\left(
\widehat{\mathbf{S}}_{t}^{_\mathrm{futu}},
\mathbf{S}_{t}^{_\mathrm{futu}}
\right),
\end{equation}
where $\mathbf{S}_{t}^{_\mathrm{mask}}$ and $\mathbf{S}_{t}^{_\mathrm{futu}}$ indicate the original latent masked representations and the target future representations, respectively. $\ell(\cdot,\cdot)$ denotes the latent space prediction loss, for which we use mean squared error (MSE). Therefore, the overall training objective is formulated as
\begin{equation}\label{eq:overall_loss}
\mathcal{L}
=
\lambda_{\mathrm{mask}}\mathcal{L}_{\mathrm{mask}}
+
\lambda_{\mathrm{futu}}\mathcal{L}_{\mathrm{futu}},
\end{equation}
where $\lambda_{\mathrm{mask}}$ and $\lambda_{\mathrm{futu}}$ are hyper-parameters.
We provide a theoretical analysis of stability and error accumulation in Appendix~\ref{appx:theory}.
The overall training algorithm is in Appendix~\ref{appx:alg}. 
\paragraph{Inference Stage.}
During inference, the relation-aware object masking and auxiliary masked-state prediction head are NOT used. Given the FULLY observed historical object representations, we apply Eq.~(\ref{eq:f_enc}) to learn the object-level memory states over the history window, and then, starting from the last historical memory state. The memory-driven future predictor performs autoregressive rollout by Eq.~(\ref{eq:f_pred}). Following the standard objective of latent world modeling, this procedure will be recursively conducted for the next prediction step until the target prediction horizon is reached.
\newcommand{\heatcell}[2]{\cellcolor{blue!#1}#2}

\section{Experiments}
To verify the effectiveness of our proposed \method, we conduct extensive experiments on visual reasoning and robotic manipulation tasks and aim to answer the following research questions: 
\textbf{RQ1} [Sec~\ref{sec:main-result}]: How does the overall performance of \method in predicting future object dynamics compare with existing predictive world-modeling baselines?
\textbf{RQ2} [Sec~\ref{sec:ablation-result}]: How do ablation studies reveal the contributions of key components in \method, particularly latent dynamic graph construction and different object masking strategies?
\textbf{RQ3} [Sec~\ref{sec:in-depth_analysis}]: How effectively can \method model temporal physical interactions?
\textbf{RQ4} [Sec~\ref{sec:in-depth_analysis}]: How robust and computationally efficient is \method in terms of hyperparameter sensitivity and running cost? More experimental results, analysis, and implementation details are listed in Appendix~\ref{appx:exps}.

\subsection{Experimental Setup}
\begin{table*}[!t]
\centering
\caption{VQA accuracy (\%) comparison on CLEVRER using VideoSAUR and SAVi encoders. \scriptsize{\textsc{WaG}$_{\mathrm{I}}$ and \textsc{WaG}$_{\mathrm{B}}$ denote instance-wise masking and batch-shared masking, respectively. $\Delta_{\mathrm{I}}$ and $\Delta_{\mathrm{B}}$ indicate the absolute improvements over C-JEPA in percentage points (pp). The best results are shown in bold, and the second-best results are underlined.}}
\vspace{-8pt}
\label{tab:vqa_comparison}
\resizebox{\textwidth}{!}{%
\begin{tabular}{lcccccccc}
\toprule
\multirow{2}{*}{Model}
& \multicolumn{1}{c}{Average}
& \multicolumn{2}{c}{Counterfactual (\%)}
& \multicolumn{2}{c}{Explanatory (\%)}
& \multicolumn{2}{c}{Predictive (\%)}
& \multirow{2}{*}{Descriptive (\%)} \\
\cmidrule(lr){3-4}
\cmidrule(lr){5-6}
\cmidrule(lr){7-8}
& per que. (\%)
& per opt. & per que.
& per opt. & per que.
& per opt. & per que.
& \\
\midrule
\multicolumn{9}{c}{\textit{VideoSAUR Encoder}} \\\midrule
OC-JEPA
& 82.79
& 79.53
& 47.68
& 92.88
& 80.58
& 86.15
& 75.04
& 89.59 \\

C-JEPA
& 89.40
& 88.67
& 68.81
& 96.62
& 90.74
& \underline{93.03}
& \underline{86.93}
& 92.84 \\ \midrule

\textbf{\textsc{WaG}$_{\mathrm{I}}$~(ours)}
& \underline{90.79} 
& \underline{90.06} 
& \underline{72.22} 
& \underline{97.79}
& \underline{93.96}
& 92.92 
& 86.81 
& \underline{93.71} \\
\rowcolor{gray!8}
\scriptsize{$\Delta_{\mathrm{I}}$ Improv.$\uparrow$}
& \scriptsize{+1.39}
& \scriptsize{+1.39}
& \scriptsize{+3.41}
& \scriptsize{+1.17}
& \scriptsize{+3.22}
& \scriptsize{-0.11}
& \scriptsize{-0.12}
& \scriptsize{+0.87} \\
\textbf{\textsc{WaG}$_{\mathrm{B}}$~(ours)}
& \textbf{91.37}
& \textbf{90.55} 
& \textbf{73.76} 
& \textbf{97.84}
& \textbf{94.13}
& \textbf{94.43} 
& \textbf{89.54} 
& \textbf{94.05} \\
\rowcolor{gray!8}
\scriptsize{$\Delta_{\mathrm{B}}$ Improv.$\uparrow$}
& \scriptsize{+1.97}
& \scriptsize{+1.88}
& \scriptsize{+4.95}
& \scriptsize{+1.22}
& \scriptsize{+3.39}
& \scriptsize{+1.40}
& \scriptsize{+2.61}
& \scriptsize{+1.21} \\
\midrule
\multicolumn{9}{c}{\textit{SAVi Encoder}} \\\midrule

OC-JEPA
& 77.28
& 76.69
& 41.10
& 91.20
& 76.04
& 83.48
& 70.51
& 84.05 \\

C-JEPA
& 83.88
& 85.16
& 60.19
& 95.34
& 87.27
& 87.46
& 77.25
& 87.81 \\
\midrule

\textbf{\textsc{WaG}$_{\mathrm{I}}$~(ours)}
 & \underline{92.34} 
 & \underline{90.98} 
 & \underline{74.62} 
 & \underline{98.25} 
 & \underline{95.18} 
 & \underline{94.63} 
 & \underline{90.05} 
 & \underline{95.06} \\
\rowcolor{gray!8}
\scriptsize{$\Delta_{\mathrm{I}}$ Improv.$\uparrow$}
& \scriptsize{+8.46}
& \scriptsize{+5.82}
& \scriptsize{+14.43}
& \scriptsize{+2.91}
& \scriptsize{+7.91}
& \scriptsize{+7.17}
& \scriptsize{+12.80}
& \scriptsize{+7.25} \\
\textbf{\textsc{WaG}$_{\mathrm{B}}$~(ours)}
& \textbf{92.72} 
& \textbf{91.19} 
& \textbf{75.45} 
& \textbf{98.35} 
& \textbf{95.35} 
& \textbf{95.84} 
& \textbf{91.99}
& \textbf{95.29} \\
\rowcolor{gray!8}
\scriptsize{$\Delta_{\mathrm{B}}$ Improv.$\uparrow$}
& \scriptsize{+8.84}
& \scriptsize{+6.03}
& \scriptsize{+15.26}
& \scriptsize{+3.01}
& \scriptsize{+8.08}
& \scriptsize{+8.38}
& \scriptsize{+14.74}
& \scriptsize{+7.48} \\

\bottomrule
\end{tabular}%
}
\vspace{-20pt}
\end{table*}
\paragraph{Tasks and Datasets.}
We evaluate the proposed \method over two tasks, i.e., \textit{visual reasoning} and \textit{robotic manipulation}.
For the \textit{visual reasoning} task, we use the CLEVRER~\citep{yi2020clevrer} dataset, a synthetic video benchmark of multiple colliding objects for physical and causal understanding in dynamic scenes, where four types of questions (i.e., counterfactual, explanatory, predictive, and descriptive) are included for the visual question-answering (VQA) task.
Each video is encoded into object-centric latent states using a frozen pretrained encoder. Following~\cite{nam2026causal}, we use VideoSAUR~\citep{zadaianchuk2023object} and SAVi~\citep{kipf2022conditional} with seven object slots. Our \method is trained on these latent states and autoregressively rolls out future object trajectories.
Following SlotFormer~\citep{wu2023slotformer}, we use ALOE~\citep{ding2021attention} to reason over predicted object trajectories and report per-question-type and average VQA accuracy.
For the \textit{robotic manipulation} task, we use the contact-rich PushT robotic manipulation benchmark~\citep{chi2025diffusion}, where an agent pushes a T-shaped block from a random pose to a target pose through sequential contact interactions.
Following~\citep{zhou2025dino,nam2026causal}, we evaluate PushT planning by success rate using six latent tokens (four object slots and two auxiliary tokens).
An episode is successful when the final block state falls within a predefined threshold of the target.

\paragraph{Baselines.} To ensure a fair comparison under consistent experimental settings, we mainly compare the proposed \method with existing state-of-the-art object-centric world models that adopt comparable latent representations and evaluation protocols. For VQA evaluation, all compared methods use the same pretrained encoder, rollout procedure, and downstream models to ensure a controlled and fair comparison. For the \textit{visual reasoning} task, we consider two state-of-the-art baseline methods: (1) OC-JEPA~\citep{nam2026causal}, and (2) Causal-JEPA (C-JEPA)~\citep{nam2026causal}. For the \textit{robotic manipulation} task, we take the following baseline models: (1) DINO-WM~\citep{zhou2025dino}, (2) DINO-WM-Reg.~\citep{darcet2024vision}, (3) OC-DINO-WM~\citep{nam2026causal}, (4) OC-JEPA, and (5) C-JEPA. All baselines start from DINOv2 embeddings~\citep {oquab2024dinov2} and differ only in predictor training. More details of all baseline methods are listed in Appendix~\ref{appx:exps}.

\subsection{Performance of \method on Visual Reasoning and Robotic Manipulation} \label{sec:main-result}
To answer \textbf{RQ1} on the \textit{visual reasoning} task, we report the VQA accuracy (\%) across four question types on the CLEVRER dataset in Table~\ref{tab:vqa_comparison}.
Overall, both variants of our proposed \method consistently achieve strong VQA performance across the VideoSAUR and SAVi encoders. In particular, \textsc{WaG}${_\mathrm{B}}$ improves the average accuracy over C-JEPA by 1.97 and 8.84 pp with VideoSAUR and SAVi encoders, respectively. Concretely, we can make the following essential observations. \textit{\underline{First}}, our proposed \method consistently improves over C-JEPA across both object-centric encoders, showing that its effectiveness does not depend on a specific pretrained encoder. 
\textit{\underline{Second}}, our proposed \method achieves the largest gains on counterfactual and predictive reasoning tasks, improving the per-question accuracy over C-JEPA by 15.26 and 14.74 pp, respectively, with the SAVi encoder. We attribute this to the proposed latent dynamic graphs and object-centric memory transition module, which explicitly models the temporal evolution of object states and relations, enabling stronger reasoning about unobserved and future dynamics.
\textit{\underline{Third}}, batch-shared masking consistently outperforms instance-wise masking across both encoders, indicating more stable learning for future prediction.

\begin{wraptable}{r}{0.5\columnwidth}
\centering
\renewcommand{\arraystretch}{0.9}
\setlength{\tabcolsep}{7pt}
\vspace{-2pt}
\caption{PushT planning success rates across different world-model token budgets.} 
\vspace{-5pt}
\label{tab:pusht_result}
\resizebox{\linewidth}{!}{%
\begin{tabular}{llc}
\toprule
\# Token $\times d$
& Model
& Success Rate (\%) \\
\midrule

\multirow{2}{*}{$196 \times 384$}
& DINO-WM
& \textbf{91.33} \\

& DINO-WM-Reg.
& 88.00 \\

\midrule

\multirow{4}{*}{$6 \times 128$}
& OC-DINO-WM (ref.)
& 60.67 \\

& OC-JEPA
& 76.00 {\scriptsize\colorbox{gray!10}{\(\uparrow\!+15.33\)}} \\

& C-JEPA
& 88.67
  {\scriptsize\colorbox{gray!10}{\(\uparrow\!+28.00\)}} \\

& \textbf{\method~(ours)}
& \underline{90.70}
  {\scriptsize\colorbox{gray!10}{\(\uparrow\!+30.03\)}} \\

\bottomrule
\end{tabular}%
}
\vspace{-10pt}
\end{wraptable}
To answer \textbf{RQ1} on the \textit{robotic manipulation} task, we compare planning performance on PushT under different world-model token budgets in Table~\ref{tab:pusht_result}. The results show that \method achieves the best success rate of 90.70\% using only $6\times128$ latent tokens, which is close to DINO-WM, achieving 91.33\% with a much larger token budget of $196\times384$. Under the same compact $6\times128$ setting, \method clearly outperforms all object-centric baselines. It exceeds JEPA-style baselines, OC-JEPA and C-JEPA, by 14.70 and 2.03 percentage points, respectively. This further demonstrates that our proposed relational structure modeling in \method can improve compact object-centric representations for downstream robotic planning.

\subsection{Ablation Study of \method on Key Components}
\label{sec:ablation-result}
\paragraph{Overall Ablation.} To answer \textbf{RQ2}, we report the overall ablation results in Table~\ref{tab:component_ablation}. It shows that all components contribute to the final performance of \method. Full WAG achieves the best average accuracy of 91.37\%, improving C-JEPA by 1.97 percentage points. Removing the memory predictor causes the largest degradation, particularly on predictive questions, where accuracy drops from 89.54\% to 85.55\%. Removing DyGraph also leads to a clear performance decrease, while Relational Masking and TGNN provide further consistent gains. These results attribute the overall improvement to the complementary effects of relation-aware structure induction and object-centric memory transition, which jointly capture evolving object relations and support future state prediction.
\begin{table*}[!h]
\centering
\begin{minipage}[t]{0.49\textwidth}
\centering
\scriptsize
\setlength{\tabcolsep}{3pt}
\vspace{-5pt}
\captionof{table}{Overall component ablation with VideoSAUR representations.}
\label{tab:component_ablation}
\vspace{-5pt}
\resizebox{\linewidth}{!}{%
\begin{tabular}{lccccc}
\toprule
Variants
& DyG
& Rel. Mask
& TGNN
& Mem. Pred.
& Avg. (\%) \\
\midrule

C-JEPA
& $\times$
& $\times$
& $\times$
& $\times$
& 89.40 \\

\midrule

w/o DyG
& $\times$
& $\times$
& $\times$
& $\checkmark$
& 90.63 \\

w/o Rel. Mask
& $\checkmark$
& $\times$
& $\checkmark$
& $\checkmark$
& 90.94 \\

w/o TGNN
& $\checkmark$
& $\checkmark$
& $\times$
& $\checkmark$
& 90.97 \\

w/o Mem. Pred.
& $\checkmark$
& $\checkmark$
& $\checkmark$
& $\times$
& 90.21 \\

\midrule

\textbf{Full \method}
& $\checkmark$
& $\checkmark$
& $\checkmark$
& $\checkmark$
& \textbf{91.37} \\

\bottomrule
\end{tabular}%
}
\end{minipage}
\hfill
\begin{minipage}[t]{0.49\textwidth}
\centering
\scriptsize
\setlength{\tabcolsep}{4pt}
\vspace{-5pt}
\captionof{table}{Comparison of different masking strategies with SAVi representations.}
\label{tab:mask_strategy}
\vspace{-5pt}
\renewcommand{\arraystretch}{1.6}
\resizebox{\linewidth}{!}{%
\begin{tabular}{lccc}
\toprule
Masking Strategy
& Avg.
& Counter.
& Predict. \\
\midrule

Random [w/o DyG.]
& 83.88
& 60.19
& 77.25 \\

\midrule

Random [w/ DyG.]
& 92.28 {\scriptsize(+8.40)}
& 73.75 {\scriptsize(+13.56)}
& 91.14 {\scriptsize(+13.89)} \\

Relational Centrality
& \textbf{92.72} {\scriptsize(+8.84)}
& \textbf{75.45} {\scriptsize(+15.26)}
& \textbf{91.99} {\scriptsize(+14.74)} \\

Temporal Dynamics
& 91.95 {\scriptsize(+8.07)}
& 73.51 {\scriptsize(+13.32)}
& 89.71 {\scriptsize(+12.46)} \\

\bottomrule
\end{tabular}%
}

\end{minipage}
\vspace{-15pt}
\end{table*}

\paragraph{In-Depth Ablation on Latent Dynamic Graph Construction.}
\begin{wrapfigure}{r}{0.4\textwidth}
    \centering
    \vspace{-8pt}
\includegraphics[width=\linewidth]{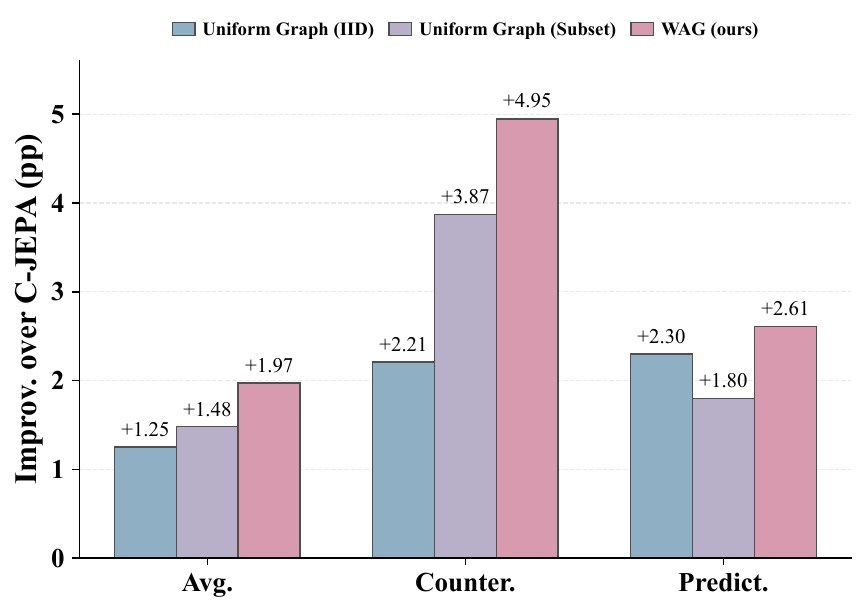}
\vspace{-20pt}
    \caption{Performance improvement of different latent dynamic graph construction strategies over C-JEPA.}
    \label{fig:graph-construction-ablation}
    \vspace{-15pt}
\end{wrapfigure}

To demonstrate the effectiveness of latent dynamic graph construction in \textbf{RQ2},
we compare our constructed KNN latent dynamic graphs in \method with two uniformly sampled graph variants. Specifically, uniform graph (IID) samples $K$ neighbors independently, allowing duplicates and self-loops, while uniform graph (subset) samples $K$ distinct neighbors without replacement and excludes self-loops.
As shown in Figure~\ref{fig:graph-construction-ablation}, our \method consistently outperforms both uniformly sampled graph variants across all CLEVRER question types.
These results indicate that the gain does not only come from introducing graph connectivity, but also from constructing informative relations in the latent space. 

\paragraph{In-Depth Ablation on Relation-Aware Object Masking.}
Table~\ref{tab:mask_strategy} compares different object masking strategies under SAVi representations in \textbf{RQ2}. Introducing dynamic graph information, i.e., random [w/ DyG.], consistently improves random masking over random [w/o DyG.], i.e., the C-JEPA setting. Among the graph-based strategies, relational centrality achieves the best performance across all reported metrics, improving the average per-question accuracy by 8.84 percentage points under SAVi representations. Temporal dynamics also provides consistent gains over the random [w/o DyG.], while relational centrality further demonstrates the benefit of selecting objects according to their relational importance.



\begin{figure}[!t]
    \centering
    \vspace{-5pt}
\includegraphics[width=1\linewidth]{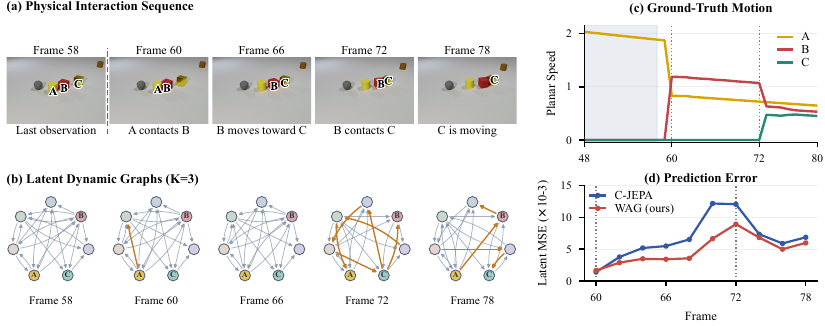}
\vspace{-20pt}
\caption{Visualization of physical interaction modeling on CLEVRER.\scriptsize{(a) Physical interaction from the last observed to future frames. (b) Latent dynamic graphs constructed by \method with $K=3$, showing evolving object relations. A/B/C correspondences are inferred. (c) Ground-truth object motion changes. (d) Latent prediction errors of C-JEPA and \method, with lower errors achieved by \method during interaction-driven transitions.}}
\vspace{-15pt}
\label{fig:case}
\end{figure}

\subsection{In-Depth Analysis of \method}\label{sec:in-depth_analysis}
\paragraph{Analysis of Physical Interaction Modeling.}

To answer \textbf{RQ3}, we present Figure~\ref{fig:case}, a CLEVRER sequence with a two-step collision chain, where A first strikes B and B subsequently strikes C within the prediction horizon. Although the model only observes the frames before these interactions, the latent graph changes when the physical relations evolve, especially around the second collision. Compared with C-JEPA, \method reduces the mean latent MSE (${\times10^{-3}}$) from 6.66 to 4.83, while lowering the peak error from 12.1 to 8.9. This suggests that the gain mainly comes from dynamic relation updates, which better capture interaction-driven state changes and future dynamics.

\begin{wrapfigure}{r}{0.49\textwidth}
    \vspace{-10pt}
    \centering
    \begin{minipage}[t]{0.48\linewidth}
        \vspace{0pt}
        \centering
        \includegraphics[width=\linewidth]{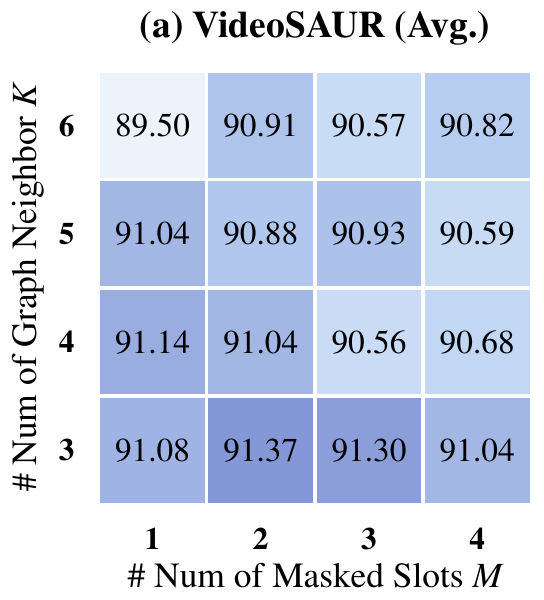}
    \end{minipage}
    \hfill
    \begin{minipage}[t]{0.48\linewidth}
        \vspace{0pt}
        \centering
        \includegraphics[width=\linewidth]{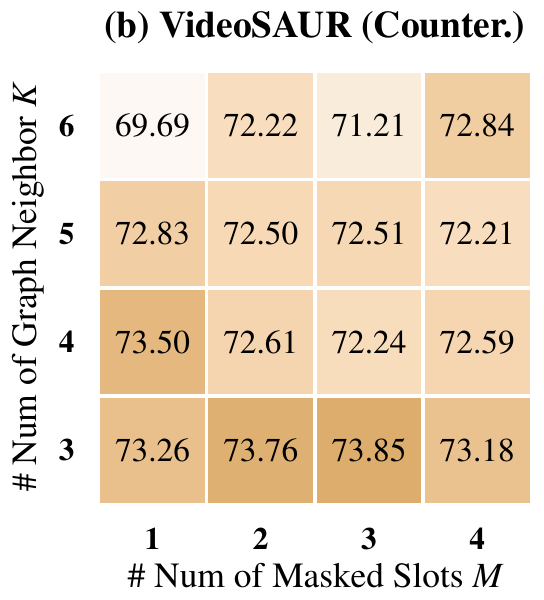}
    \end{minipage}
    \vspace{-5pt}
    \caption{Analysis of the number of masked objects $M$ and graph neighborhood size $K$.}
    \label{fig:mk_sensitivity}
    \vspace{-8pt}
\end{wrapfigure}
\paragraph{Analysis of the Number of Masked Objects $M$ and Graph Neighborhood Size $K$.}
Figure~\ref{fig:mk_sensitivity} studies the sensitivity of \method to the number of masked objects $M$ and graph neighborhood size $K$ in \textbf{RQ4}. Overall, the performance remains relatively stable across different parameter combinations. Moderate neighborhood sizes ($K{=}3$) generally achieve stronger performance, while the preferred masking number varies slightly across representations and evaluation metrics. In particular, VideoSAUR achieves its highest average accuracy of 91.37\% at $K{=}3, M{=}2$. More results are in Appendix~\ref{appx:exps}.
\paragraph{Analysis of Running Costs.} 
For \textbf{RQ4}, we report the training time and peak GPU memory. Our proposed \method is substantially more efficient than C-JEPA and OC-JEPA, requiring only about 143 seconds per epoch compared with 341.4 seconds for C-JEPA, i.e., approximately \(2.4\times\) faster training. It also reduces peak GPU memory from 6712 MiB to 410 MiB, achieving up to a \(16.4\times\) reduction. More detailed running costs, including GFLOPs, are provided in Appendix~\ref{appx:exps}.

\section{Conclusion}
In this work, we introduce \textbf{\underline{\textsc{W}}}orld-\textbf{\underline{\textsc{a}}}s-\textbf{\underline{\textsc{G}}}raph (\textbf{\method}), a graph-based object-centric world model that explicitly captures object relations and their temporal evolution in latent space. \method integrates relation-aware structure induction with object-centric memory transition, allowing relational information and historical object states to jointly support future prediction. 
Experiments on visual reasoning and robotic manipulation tasks demonstrate the consistent improvements of our proposed \method.  In-depth ablation studies and analysis, along with physical interaction analysis, further highlight the effectiveness of representing the latent world as dynamic graphs. More broadly, \method provides a general framework for connecting object-centric representation learning with relational graph modeling, offering a step toward more structured and interaction-aware world models.

\subsection*{AI use statement}
In this work, we used generative AI tools for editing and improving the readability of the manuscript, refining the presentation of scientific figures, and supporting software code development and debugging. The research ideas, proposed methodology, experimental design, and interpretation of the experimental results were developed and conducted by the authors. All AI-assisted text was reviewed and revised by the authors. No AI-generated content was directly adopted without author verification. We take responsibility for the final content of this work, including text, claims, code, figures, and other artifacts produced with the aid of generative AI.

\subsection*{Ethics Statement}
This work does not involve human subjects, personal or sensitive data, or other ethical concerns requiring specific consideration.

\subsection*{Reproducibility Statement}
We provide the main architectural details and learning objectives of \method in the Method section. The datasets, evaluation protocols, baseline settings, and partial implementation details are described in the Experiment section. Additional hyperparameter settings, model configurations, and implementation details required for reproducing the reported results are provided in the Appendix.


\bibliography{iclr2027_conference}
\bibliographystyle{iclr2027_conference}
\newpage
\appendix
\setcounter{table}{0}
\setcounter{figure}{0}

\renewcommand{\thetable}{A\arabic{table}}
\renewcommand{\thefigure}{A\arabic{figure}}
\section{Appendix}
This is the Appendix of the submission: \textbf{World-As-Graph: Relational World Modeling Through Latent Space Graphs}.
In this appendix, we include more details of: (1) Related work, (2) Theoretical analysis, (3) Algorithm, and (4) Experimental details and comprehensive results.

\subsection{Related Work}\label{appx:related_work}
\paragraph{Joint-Embedding Predictive Architectures.} As a common architecture for self-supervised representation learning, the joint-embedding predictive architecture (JEPA)~\citep{lecun2022path,assran2023self,klindt2026does} aims to train an encoder and a predictor based on observations to learn informative representations of inputs in a latent space, so that it could forecast future world states from context and actions by capturing the input relationships~\citep{klindt2026does,bai2026temporal,saito2025point,tuncay2025audio}.
Based on this architecture, JEPA has been applied on different modalities, such as Point-JEPA~\citep{saito2025point} on point clouds and Audio-JEPA~\citep{tuncay2025audio} on audio signals. LeJEPA~\citep{balestriero2025lejepa} further studies that the latent prediction objectives can be trained in a scalable manner without relying on heuristics. Moreover, different research have been studied for dynamic scenes, the predictor is required to forecast the latent states of future observations. 
Temporal-Distance JEPA~\citep{bai2026temporal} shapes the latent space with a temporal distance, where these trained representations are shaped to be meaningful for downstream planning. Chain-of-World predicts~\citep{yang2026chain} the future through a chain of latent motion states. Causal ideas have also been introduced into the objective, Causal-JEPA~\citep{nam2026causal} masks the latent states, forcing each masked object to be predicted from the remaining objects, although which objects are masked is still decided at random. 
\paragraph{Graph World Models.}
Recent graph world models mainly explore \textit{world modeling for graphs}, where graph-structured states are treated as the environment representation for prediction, simulation, and planning~\citep{liu2026graph,feng2025graph}. For example, GWM~\citep{feng2025graph} provides a unified framework for modeling graph-structured and multimodal states through message passing and action nodes. More recent studies further investigate rollout errors over graph trajectories~\citep{song2026understanding} and incorporate explicit structural mechanisms and constraints into graph-based dynamics~\citep{wang2026structural}. In these methods, graph structure mainly serves as the state space on which world dynamics are modeled. A related line of work introduces graph structures into object-centric world models to represent interactions among visual entities. C-SWM~\citep{kipf2020contrastive} represents object states and their relations through a graph neural network, while FIOC-WM~\citep{feng2026learning} starts learning interaction structures between object-centric representations.
Different from these studies, our work focuses on \textit{graphs for world modeling}: latent dynamic graphs are introduced as relational inductive biases to guide object-centric predictive representation learning. In particular, \method uses evolving graph structures not only to represent object interactions, but also to guide relational masking and object-level memory transition, encouraging future prediction to exploit both inter-object dependencies and temporal dynamics.

\paragraph{Dynamic Graph Representation Learning.} 
Dynamic graphs can be widely applied in many real-world systems, where the graph topology and node attributes keep changing with time~\citep{feng2025comprehensive, zheng2025survey}. The key to dynamic graph representation learning is to capture the relations and structure across different time steps. In this context, dynamic graph neural networks (DGNNs) become powerful tools to learn temporal entity interactions and representations in the real world.
Typically, TGAT~\citep{xu2020inductive} combines time encoding with temporal attention over neighbors. TGN~\citep{rossi2020temporal} chooses to maintain a recurrent memory state for each graph node and keep updating the state with each interaction. NeurTWs~\citep{jin2022neural} effectively capture each temporal node while preserving temporal constraints by conducting spatiotemporal-biased random walks. More recent models simplify the architecture. GraphMixer~\citep{cong2023we} shows that a simple MLP-based encoder is already competitive, while DyGFormer~\citep{yu2023towards} learns from the neighbor sequence of each node with a transformer. GraphSSM~\citep{li2024state} models the evolution of temporal graphs with a state space formulation. MaskDGNN~\citep{he2025maskdgnn} masks the temporal edges in a self-supervised manner. This line of research is closely related to the temporal GNN component in \method, while our goal is not general-purpose dynamic graph representation learning.
\begin{algorithm}
\caption{\method Training}
\label{alg:wag-train}
\small
\begin{algorithmic}[1]
\REQUIRE Slot sequence $\{\mathbf{S}_t\}_{t\in\mathcal{T}}$ from $f_{\Phi^*}$, $\mathcal{T}=\mathcal{T}_{\text{hist}}\cup\mathcal{T}_{\text{pred}}$; history/horizon lengths $T_h, T_p$; graph degree $K$; mask budget $M$; 

masking policy $\pi_{\text{mask}} \in \{\text{Relational Centrality}, \text{Temporal Dynamics}\}$; loss weights $\lambda_{\mathrm{mask}}, \lambda_{\mathrm{futu}}$; 
\ENSURE Learned parameters $(\theta, \omega)$ of \method (best checkpoint by validation loss)
\REPEAT
    \STATE Sample $\{\mathbf{S}_t\}_{t\in\mathcal{T}}$; let $t_0 = t - T_h + 1$
    \STATE $G_t \leftarrow \Gamma_K^{\text{sim}}(\mathbf{S}_t)$ for each $t \in \mathcal{T}_{\text{hist}}$ Eq.~(\ref{eq:dynamic_graph})
    \STATE Compute $c_i$ Eq.~(\ref{eq:degree_mask}) or $d_i$ Eq.~(\ref{eq:temporal_mask}) per $\pi_{\text{mask}}$; $\mathcal{I}_{\text{mask}} \leftarrow \arg\text{Top-}M$
    \STATE $G_t^\dagger \leftarrow (\mathcal{V}_t, \mathcal{E}_t, \mathbf{S}_t^\dagger)$ for $t \in \mathcal{T}_{\text{hist}}\setminus\{t_0\}$, masking $\mathbf{s}_t^i,\, i\in\mathcal{I}_{\text{mask}}$; keep $G_{t_0}$ fully visible
    \STATE $\mathbf{H}_{t_0} \leftarrow \mathbf{S}_{t_0}$
    \FOR{$t \in \mathcal{T}_{\text{hist}} \setminus \{t_0\}$}
        \STATE $\mathbf{H}_t \leftarrow \text{MemGRU}_{\theta_m}\big(\text{TGNN}_{\theta_g}(G_t^\dagger), \mathbf{H}_{t-1}\big)$ Eq.~(\ref{eq:f_enc})
        \STATE $\widehat{\mathbf{S}}_t^{\text{mask}} \leftarrow g_{\text{mask}}^\epsilon(\mathbf{H}_t, \mathcal{I}_{\text{mask}})$ Eq.~(\ref{eq:mask_pred_representation})
    \ENDFOR
    \FOR{$t \in \mathcal{T}_{\text{pred}}$}
        \STATE $\widehat{G}_{t-1} \leftarrow \Gamma_K^{\text{sim}}(\mathbf{H}_{t-1})$; \; $\mathbf{H}_t \leftarrow f_{\text{pred}}^\omega(\widehat{G}_{t-1}, \mathbf{H}_{t-1})$; \; $\widehat{\mathbf{S}}_t^{\text{futu}} \leftarrow g_{\text{proj}}^\eta(\mathbf{H}_t)$ Eq.~(\ref{eq:f_pred})
    \ENDFOR
    \STATE $\mathcal{L}_{\mathrm{mask}}, \mathcal{L}_{\mathrm{futu}} \leftarrow$ Eq.~(\ref{eq:lmask_lfutu}); \quad $\mathcal{L}(\theta,\omega) \leftarrow \lambda_{\mathrm{mask}}\mathcal{L}_{\mathrm{mask}} + \lambda_{\mathrm{futu}}\mathcal{L}_{\mathrm{futu}}$ Eq.~(\ref{eq:overall_loss})
    \STATE $(\theta, \omega) \leftarrow \text{AdamW}\big((\theta,\omega), \nabla_{\theta,\omega} \mathcal{L}\big)$
    \IF{$\mathcal{L}_{\text{val}} < \mathcal{L}_{\text{best}}$}
        \STATE $(\theta_{\text{best}}, \omega_{\text{best}}) \leftarrow (\theta, \omega)$; \; $\mathcal{L}_{\text{best}} \leftarrow \mathcal{L}_{\text{val}}$
    \ENDIF
\UNTIL converged
\RETURN $(\theta_{\text{best}}, \omega_{\text{best}})$
\end{algorithmic}
\end{algorithm}

\subsection{Theoretical Analysis}\label{appx:theory}
During autoregressive rollout, \method constructs future latent graphs from evolving object-level memory states without access to future observations. Consequently, even small memory errors can change neighbor selection, alter relational message passing, and affect subsequent predictions. This coupling raises a central question:

\begin{tcolorbox}[
    enhanced,
    frame hidden,
    boxrule=0pt,
    sharp corners,
    colback=black!8,
    borderline west={1.25pt}{0pt}{black!80},
    boxsep=0pt,
    left=7pt,
    right=7pt,
    top=4pt,
    bottom=4pt,
    before skip=6pt,
    after skip=6pt,
    fontupper=\normalsize\itshape
]
Under what conditions do memory perturbations preserve KNN
neighborhoods relative to a reference trajectory, and how do
these errors accumulate across rollout steps?
\end{tcolorbox}

To answer this question, we first characterize neighborhood stability through the cosine-similarity margin at the KNN selection boundary. We then derive a memory-error bound that connects this margin condition with transition sensitivity and single-step prediction residuals.

We analyze how memory perturbations affect $\Gamma_K^{\mathrm{sim}}$ and $f_{\mathrm{pred}}^\omega$ in Eq.~(\ref{eq:f_pred}). Let  $\mathbf{H}_t$ denote the rollout memories and $\mathbf{H}_t^\star$ a reference memory trajectory for analysis, with all node norms bounded below by $r>0$. Therefore, we define
\begin{equation}
e_t=\|\mathbf{H}_t-\mathbf{H}_t^\star\|_{\max},
\qquad
\|\mathbf{H}\|_{\max}=\max_i\|\mathbf{h}^i\|_2.
\label{eq:memory_error}
\end{equation}
For $1\le K<N-1$, let $w_{i,(k)}^{t,\star}$ denote the $k$-th largest value of $\cos(\mathbf{h}_t^{i,\star},\mathbf{h}_t^{j,\star})$ over $j\ne i$. We define the neighborhood margin as
\begin{equation}
\gamma_t=\min_i\left(w_{i,(K)}^{t,\star}-w_{i,(K+1)}^{t,\star}\right).
\end{equation}
Normalized node discrepancies are at most $2e_t/r$, so each cosine similarity changes by at most $4e_t/r$. Therefore, $e_t<r\gamma_t/8$ preserves neighbor sets between predicted and reference graphs, without requiring constant topology over time or unchanged edge weights.
Assume the memory update in Eq.~(\ref{eq:f_pred}) is uniformly $L$-Lipschitz under $\|\cdot\|_{\max}$ for fixed neighbor indices and recomputed cosine weights, with reference residual as:
\begin{equation}
\left\|
\mathbf{H}_{t+1}^\star-
f_{\mathrm{pred}}^\omega
\left(\Gamma_K^{\mathrm{sim}}(\mathbf{H}_t^\star),\mathbf{H}_t^\star\right)
\right\|_{\max}
\le\delta.
\end{equation}
Let $T=\max\mathcal{T}_{\mathrm{hist}}$. If the margin condition holds throughout the rollout, then, for $1\le h\le T_p$, we have:
\begin{equation}
e_{t+1}\le Le_t+\delta,
\qquad
e_{T+h}\le L^he_T+\delta\sum_{j=0}^{h-1}L^j.
\end{equation}
The triangle inequality yields the one-step bound; induction gives the multi-step bound. Thus, neighborhood separation prevents perturbation-induced rewiring, while transition sensitivity controls error accumulation. For $L<1$, the bound remains uniform over the horizon. These are sufficient analytical conditions, not guarantees provided by recurrent memory alone.

\subsection{Algorithms} \label{appx:alg}
Algorithm~\ref{alg:wag-train} summarizes the overall training procedure of our proposed \method.

\subsection{Experimental Details and Comprehensive Results}\label{appx:exps}

\paragraph{Dataset Overview.}
\begin{wraptable}{r}{0.6\columnwidth}
\vspace{-8pt}
\centering
\caption{Dataset statistics.}
\vspace{-8pt}
\label{appx:dataset-comparison}
\setlength{\tabcolsep}{3.5pt}
\renewcommand{\arraystretch}{1}

\resizebox{\linewidth}{!}{%
\begin{tabular}{lll}
\toprule
\textbf{Property} & \textbf{PushT} & \textbf{CLEVRER} \\
\midrule
Train split      & 18{,}685 trajectories & 10{,}000 videos \\
Val / test split & 21 trajectories & 5{,}000 / 5{,}000 videos \\
Object slots     & 4 & 7 \\
Task type        & 2D pushing manipulation & Video QA / causal reasoning \\
Downstream head  & CEM planner & ALOE-style \\
Evaluation metric & Success rate & QA accuracy \\
\bottomrule
\end{tabular}%
}

\vspace{-8pt}
\end{wraptable}
PushT~\citep{chi2025diffusion} and CLEVRER ~\citep{yi2020clevrer} represent two distinct downstream tasks of \method used in this paper. CLEVRER is a discrete video question-answering(VQA) and causal reasoning task, where an ALOE-style VQA head is used to classify four types of questions: \textit{counterfactual questions}, asking how the scene would evolve if a certain object is removed; \textit{explanatory questions}, asking which observed event is responsible for a given event; \textit{predictive questions}, asking which event will happen after the video ends; \textit{descriptive questions}, asking about the objects and the events in the observed frames. Performance on CLEVRER is evaluated using question-answering accuracy. 

In contrast, PushT is a continuous control and planning task, which is inferred with the Cross-Entropy Method (CEM) over the learned world model, and performance is measured by task success rate. At every planning step, CEM samples a population of candidate action sequences and scores them by the learned world model's predicted cost to the goal. An episode is deemed successful if, within this budget, the agent-block configuration matches the goal state within a position error of 20 pixels and an orientation error of $\pi/9$ radians. Performance on PushT is evaluated using the success rate. 
The dataset statistics are summarized in Table~\ref{appx:dataset-comparison}.

\paragraph{Baselines.}
For the \textit{visual reasoning} task, specifically, we consider two baseline methods: (1) OC-JEPA~\citep{nam2026causal}, which predicts the future object slots from a fully observed history; and (2) Causal-JEPA (C-JEPA)~\citep{nam2026causal}, which additionally masks a subset of object slots within the observed history and jointly trains the model to reconstruct the masked slots while predicting the future object slots. For VQA evaluation, all compared methods use the same pretrained encoder, rollout procedure, and downstream models to ensure a controlled and fair comparison. 

For the \textit{robotic manipulation} task, we take the following baseline models: (1) DINO-WM~\citep{zhou2025dino}, an autoregressive predictor that predicts over patch-level representations; (2) DINO-WM-Reg.~\citep{darcet2024vision}, a variant that uses a DINOv2 backbone while retaining the same patch-based predictor, thereby isolating the effect of the representation; (3) OC-DINO-WM~\citep{nam2026causal}, which replaces DINO-WM's patch embeddings with object-centric slots; (5) OC-JEPA and C-JEPA~\citep{nam2026causal}, which uses the same object-slot masking objectives introduced above and evaluated here on planning. To maintain comparability, all baselines start from DINOv2 embeddings~\citep {oquab2024dinov2} and differ only in predictor training.

\begin{table*}[!t]
\centering
\begin{minipage}[t]{0.49\textwidth}
\centering
\caption{Implementation details and hyperparameters for the CLEVRER experiments.}
\label{appx:clevrer-hparams}
\setlength{\tabcolsep}{4pt}
\resizebox{\linewidth}{!}{%
\begin{tabular}{ccl}
\toprule
Symbol & Value & Description \\
\midrule

\multicolumn{3}{l}{\textit{\method{} hyperparameters}} \\
$T_h$        & 6                & history window length \\
$T_p$        & 10               & prediction horizon \\
$K$          & 3/4              & DyG Builder neighbors \\
$M$          & 2/4              & mask budget \\
$\Delta_{\text{frame}}$ & 2     & frame skip \\
$\lambda_{\mathrm{mask}}$ & 1     & masked-history loss weight \\
$\lambda_{\mathrm{futu}}$ & 0.25/0.5 & future-prediction loss weight \\
$\alpha_{\text{pred}}$  & $5\times10^{-4}$ & predictor learning rate \\
$n_{\text{batch}}$      & 256   & batch size \\
$E$                      & 60    & predictor training epochs \\
                         & best validation loss & checkpoint selection \\
\midrule

\multicolumn{3}{l}{\textit{Downstream VQA (ALOE)}} \\
$E_{\text{ALOE}}$       & 400              & ALOE training epochs \\
$\alpha_{\text{ALOE}}$  & $1\times10^{-3}$ & ALOE learning rate \\
                         & fp16             & precision \\
                         & 0                & random seed \\
\bottomrule
\end{tabular}}
\end{minipage}
\hfill
\begin{minipage}[t]{0.49\textwidth}
\centering
\caption{Implementation details and hyperparameters for the PushT experiments.}
\label{appx:pusht-hparams}
\setlength{\tabcolsep}{4pt}
\resizebox{\linewidth}{!}{%
\begin{tabular}{ccl}
\toprule
Symbol & Value & Description \\
\midrule

\multicolumn{3}{l}{\textit{\method{} hyperparameters}} \\
$T_h$        & 3                & history window length \\
$T_p$        & 3                & prediction horizon \\
$K$          & 2                & DyG Builder neighbors \\
$M$          & 1                & mask budget \\
$\Delta_{\text{frame}}$ & 5     & frame skip \\
$D_{\text{proprio}}$ & 128      & proprioception embedding dim \\
$D_{\text{act}}$     & 128      & action embedding dim \\

$\alpha_{\text{pred}}$    & $5\times10^{-4}$ & predictor learning rate \\
$\alpha_{\text{proprio}}$ & $1\times10^{-4}$ & proprioception encoder learning rate \\
$\alpha_{\text{act}}$     & $5\times10^{-4}$ & action encoder learning rate \\
$n_{\text{batch}}$        & 256  & batch size \\
$E$                       & 30   & predictor training epochs \\
\midrule

\multicolumn{3}{l}{\textit{Planning}} \\
$T_h$                     & 3            & history size \\
$n_{\text{seed}}$         & $\{0,1,2\}$  & evaluation seeds \\
\bottomrule
\end{tabular}
}
\end{minipage}

\end{table*}
\begin{table}[!t]
\centering
\caption{Comparison of masking strategies for \textbf{\textsc{WaG}$_{\mathrm{I}}$} vs.\ \textbf{\textsc{WaG}$_{\mathrm{B}}$} on the VideoSAUR encoder (all runs at $\lambda_{\mathrm{futu}}=1$) and $\lambda_{\mathrm{mask}}=1$.}
\setlength{\tabcolsep}{10pt}
\label{appx:wag_i_vs_b_masking}
\resizebox{\linewidth}{!}{%
\begin{tabular}{llc cc cc cc}
\toprule
\multirow{2}{*}{Masking} & \multirow{2}{*}{Model} & \multirow{2}{*}{\shortstack{Average \\ per que.\ (\%)}} & \multicolumn{2}{c}{Counterfactual (\%)} & \multicolumn{2}{c}{Explanatory (\%)} & \multicolumn{2}{c}{Predictive (\%)} \\
\cmidrule(lr){4-5} \cmidrule(lr){6-7} \cmidrule(lr){8-9}
 & & & per opt. & per que. & per opt. & per que. & per opt. & per que. \\
\midrule
\multirow{3}{*}{Random}
 & \textsc{WaG}$_{\mathrm{I}}$ & 90.94 & 90.20 & 72.83 & 97.69 & 93.54 & 93.97 & 88.75 \\
 & \textsc{WaG}$_{\mathrm{B}}$ & \textbf{91.14} & 90.31 & 72.76 & 97.71 & 93.64 & \textbf{94.21} & \textbf{89.34} \\
 & $\Delta$ (B$-$I) & {+0.20} & +0.11 & $-$0.07 & +0.02 & +0.10 & {+0.24} & {+0.59} \\
\midrule
\multirow{3}{*}{Relational Centrality}
 & \textsc{WaG}$_{\mathrm{I}}$ & \textbf{90.79} & 90.06 & \textbf{72.22} & \textbf{97.79} & \textbf{93.96} & 92.92 & 86.81 \\
 & \textsc{WaG}$_{\mathrm{B}}$ & 90.85 & 89.96 & 72.20 & 97.31 & 92.80 & \textbf{93.52} & \textbf{87.94} \\
 & $\Delta$ (B$-$I) & +0.06 & $-$0.10 & $-$0.02 & $-$0.48 & $-$1.16 & +0.60 & +1.13 \\
\midrule
\multirow{3}{*}{Temporal Dynamics}
 & \textsc{WaG}$_{\mathrm{I}}$ & 90.47 & 89.14 & 70.16 & 96.95 & 91.74 & 93.41 & 87.71 \\
 & \textsc{WaG}$_{\mathrm{B}}$ & \textbf{90.94} & \textbf{90.23} & \textbf{72.51} & \textbf{97.55} & \textbf{93.31} & \textbf{93.86} & \textbf{88.59} \\
 & $\Delta$ (B$-$I) & {+0.47} & {+1.09} & {+2.35} & {+0.60} & {+1.57} & {+0.45} & {+0.88} \\
\bottomrule
\end{tabular}%
}
\end{table}
\paragraph{Implementation Details.}
For visual reasoning over the CLEVRER dataset and robotic manipulation over the PushT dataset, \method is trained on a pretrained object-centric encoder. VideoSAUR is used for both datasets, while SAVi is additionally included for comparison on CLEVRER. All models are optimized with AdamW. For downstream evaluation, we select the checkpoint with the lowest validation loss rather than the checkpoint from the final training iteration. In the following, we provide detailed parameter settings for each experimental component. Unless otherwise specified, all experiments follow the default configuration in Table~\ref{appx:clevrer-hparams} and Table~\ref{appx:pusht-hparams}.

For PushT, the four object slots are augmented with two auxiliary tokens encoding the proprioceptive state and action, respectively. The state and action are separately projected into the same \(D\)-dimensional latent space as the object slots and appended to the object representations, resulting in six latent tokens in total. During CEM planning, each candidate action is encoded as the action token to condition the corresponding future transition, while the proprioception token provides the current robot state.

In Table~\ref{tab:vqa_comparison} of the main submission, we report the best-performing results for \textsc{WaG}$_{\mathrm{I}}$ and \textsc{WaG}$_{\mathrm{B}}$ when using Relational Centrality as the masking mechanism. For VideoSAUR, we set $\lambda_{\mathrm{futu}}=0.25$; the reported \textsc{WaG}$_{\mathrm{I}}$ results use $K{=}4$ and $M{=}4$, while \textsc{WaG}$_{\mathrm{B}}$ uses $K{=}3$ and $M{=}2$. For SAVi, we set $\lambda_{\mathrm{futu}}=0.5$; \textsc{WaG}$_{\mathrm{B}}$ uses $K{=}4$ and $M{=}2$, whereas the reported \textsc{WaG}$_{\mathrm{I}}$ results use $K{=}4$ and $M{=}4$. ALOE is trained for 400 epochs. When analyzing the effect of $\lambda_{\mathrm{futu}}$ on performance, we trained ALOE for only 100 epochs as a reference due to computational constraints.
All experiments are run on a single NVIDIA H20 GPU.
\begin{table}[!t]
\centering
\caption{Overall ablation study of key components in our proposed \method with VideoSAUR encoder. \scriptsize{Counter. and Predict. denote per-question accuracy for counterfactual and predictive questions, respectively.}}
\label{appx:component_ablation}
\resizebox{\linewidth}{!}{
\begin{tabular}{lccccccccc}
\toprule
& \multicolumn{2}{c}{
    \begin{tabular}{@{}c@{}}
    \textbf{\small{Relation-Aware}} \\
    \textbf{\small{Structure Induction}}
    \end{tabular}
}
& \multicolumn{2}{c}{
    \begin{tabular}{@{}c@{}}
    \textbf{\small{Object-Centric}} \\
    \textbf{\small{Memory Transition}}
    \end{tabular}
}
& \multicolumn{5}{c}{Performance (\%) $\uparrow$} \\
\cmidrule(lr){2-3}
\cmidrule(lr){4-5}
\cmidrule(lr){6-9}

Variants
& DyGraph
& Rel. Mask
& TGNN 
& Mem. Pred.
& Avg. per que.
& Counter. 
& Explan.
& Predict.
& Descrip. \\
\midrule

C-JEPA
& $\times$ 
& $\times$ 
& $\times$ 
& $\times$
& 89.40 
& 68.81 
& 90.74
& 86.93 
& 92.84  \\

\midrule

w/o DyGraph
& $\times$ 
& $\times$ 
& $\times$ 
& $\checkmark$
& 90.63 
& 72.68 
& 93.38 
& 89.15 
& 93.36  \\

w/o Rel. Mask
& $\checkmark$ 
& $\times$ 
& $\checkmark$ 
& $\checkmark$
& 90.94 
& 72.83 
& 93.54
& 88.75 
& 93.75  \\

w/o TGNN
& $\checkmark$ 
& $\checkmark$ 
& $\times$ 
& $\checkmark$
& 90.97 
& 73.42 
& 93.52
& 90.33 
& 93.59  \\

w/o Mem. Pred.
& $\checkmark$ 
& $\checkmark$ 
& $\checkmark$ 
& $\times$
& 90.21 
& 71.50 
& 92.46
& 85.55 
& 93.34  \\

\midrule

\textbf{Full \method (ours)}
& $\checkmark$ 
& $\checkmark$ 
& $\checkmark$ 
& $\checkmark$
& \textbf{91.37} 
& \textbf{73.76} 
& \textbf{94.13} 
& \textbf{89.54} 
& \textbf{94.05}\\
\bottomrule
\end{tabular}
}
\end{table}
\begin{table*}[!t]
\centering
\caption{Comparison of different object masking strategies under VideoSAUR and SAVi representations using per-question accuracy. 
\scriptsize{Values in parentheses denote absolute improvements over Random [w/o DyG.] (i.e., C-JEPA) in percentage points, where w/o DyG. and w/ DyG. indicate without and with latent dynamic graph construction, respectively.}}
\label{appx:mask_strategy}
\resizebox{\textwidth}{!}{%
\begin{tabular}{lcccccc}
\toprule
\multirow{2}{*}{Masking Strategy}
& \multicolumn{3}{c}{\textit{VideoSAUR Encoder}}
& \multicolumn{3}{c}{\textit{SAVi Encoder}} \\
\cmidrule(lr){2-4}
\cmidrule(lr){5-7}

& Avg. per. que. (\%)
& Counter. (\%)
& Predict. (\%)
& Avg. per. que. (\%)
& Counter. (\%)
& Predict. (\%) \\
\midrule

Random [w/o DyG.]
& 89.40
& 68.81
& 86.93
& 83.88
& 60.19
& 77.25 \\

\midrule

Random [w/ DyG.]
& 90.94 {\scriptsize(+1.54)}
& 72.83 {\scriptsize(+4.02)}
& 88.75 {\scriptsize(+1.82)}
& 92.28 {\scriptsize(+8.40)}
& 73.75 {\scriptsize(+13.56)}
& 91.14 {\scriptsize(+13.89)} \\

Relational Centrality
& 91.37 {\scriptsize(+1.97)}
& 73.76 {\scriptsize(+4.95)}
& 89.54 {\scriptsize(+2.61)}
& 92.72 {\scriptsize(+8.84)}
& 75.45 {\scriptsize(+15.26)}
& 91.99 {\scriptsize(+14.74)} \\

Temporal Dynamics
& 90.47 {\scriptsize(+1.07)}
& 70.16 {\scriptsize(+1.35)}
& 87.71 {\scriptsize(+0.78)}
& 91.95 {\scriptsize(+8.07)}
& 73.51 {\scriptsize(+13.32)}
& 89.71 {\scriptsize(+12.46)} \\

\bottomrule
\end{tabular}%
}
\end{table*}

\begin{table*}[t]
\centering
\caption{VQA accuracy comparison of different random-graph sampling strategies and \textsc{WaG}$_{\mathrm{B}}$ using the VideoSAUR encoder.}
\label{appx:random_graph_comparison}
\resizebox{\textwidth}{!}{%
\begin{tabular}{lcccccccc}
\toprule
\multirow{2}{*}{Model}
& \multicolumn{1}{c}{Average}
& \multicolumn{2}{c}{Counterfactual (\%)}
& \multicolumn{2}{c}{Explanatory (\%)}
& \multicolumn{2}{c}{Predictive (\%)}
& \multirow{2}{*}{Descriptive (\%)} \\
\cmidrule(lr){3-4}
\cmidrule(lr){5-6}
\cmidrule(lr){7-8}
& per que. (\%)
& per opt. & per que.
& per opt. & per que.
& per opt. & per que.
& \\
\midrule

C-JEPA
& 89.40
& 88.67
& 68.81
& 96.62
& 90.74
& 93.03
& 86.93
& 92.84 \\

\midrule

Random Graph (IID)
& 90.65
& 89.52
& 71.02
& 97.28
& 92.60
& 94.24
& 89.23
& 93.77 \\

Random Graph (Subset)
& 90.88
& 90.24
& 72.68
& 97.55
& 93.27
& 93.94
& 88.73
& 93.74 \\

\midrule

\textbf{\textsc{WaG}$_{\mathrm{B}}$ (ours)}
& \textbf{91.37}
& \textbf{90.55}
& \textbf{73.76}
& \textbf{97.84}
& \textbf{94.13}
& \textbf{94.43}
& \textbf{89.54}
& \textbf{94.05} \\

\bottomrule
\end{tabular}%
}
\end{table*}

\paragraph{Instance-Wise vs. Batch-Shared Masking.}
Under identical parameter settings, we also evaluate the effects of instance-wise masking and batch-shared masking across different masking strategies, with the results shown in Table~\ref{appx:wag_i_vs_b_masking}. 
For the random mask strategy, \textsc{WaG}$_{\mathrm{B}}$ also achieves small but consistent improvements over the instance-wise masking variant \textsc{WaG}$_{\mathrm{I}}$, although the gains are substantially small.

In contrast, \textsc{WaG}$_{\mathrm{B}}$ shows the clearest and most consistent advantage under the temporal dynamics strategy. In particular, it improves counterfactual per-question accuracy by 2.35 points and explanatory per-question accuracy by 1.57 points, and only a marginal decrease of 0.05 on the descriptive category. This suggests that batch-shared masking, as used in \textsc{WaG}$_{\mathrm{B}}$, is particularly effective when temporal changes drive the masking strategy.

Meanwhile, the results under relational centrality indicate that the relative benefit of batch-shared versus instance-wise masking depends on the masking strategy. Although \textsc{WaG}$_{\mathrm{B}}$ achieves a higher overall score by 0.06 points, \textsc{WaG}$_{\mathrm{I}}$ performs substantially better on explanatory per-question accuracy, with a gain of 1.16 points. This suggests that, unlike temporal-change masking, degree-based masking does not consistently benefit from batch-shared masking.

\paragraph{Ablation on Graph Connectivity and MLP Capacity.}
\begin{wrapfigure}{r}{0.48\columnwidth}
    \centering
\includegraphics[width=0.7\linewidth]{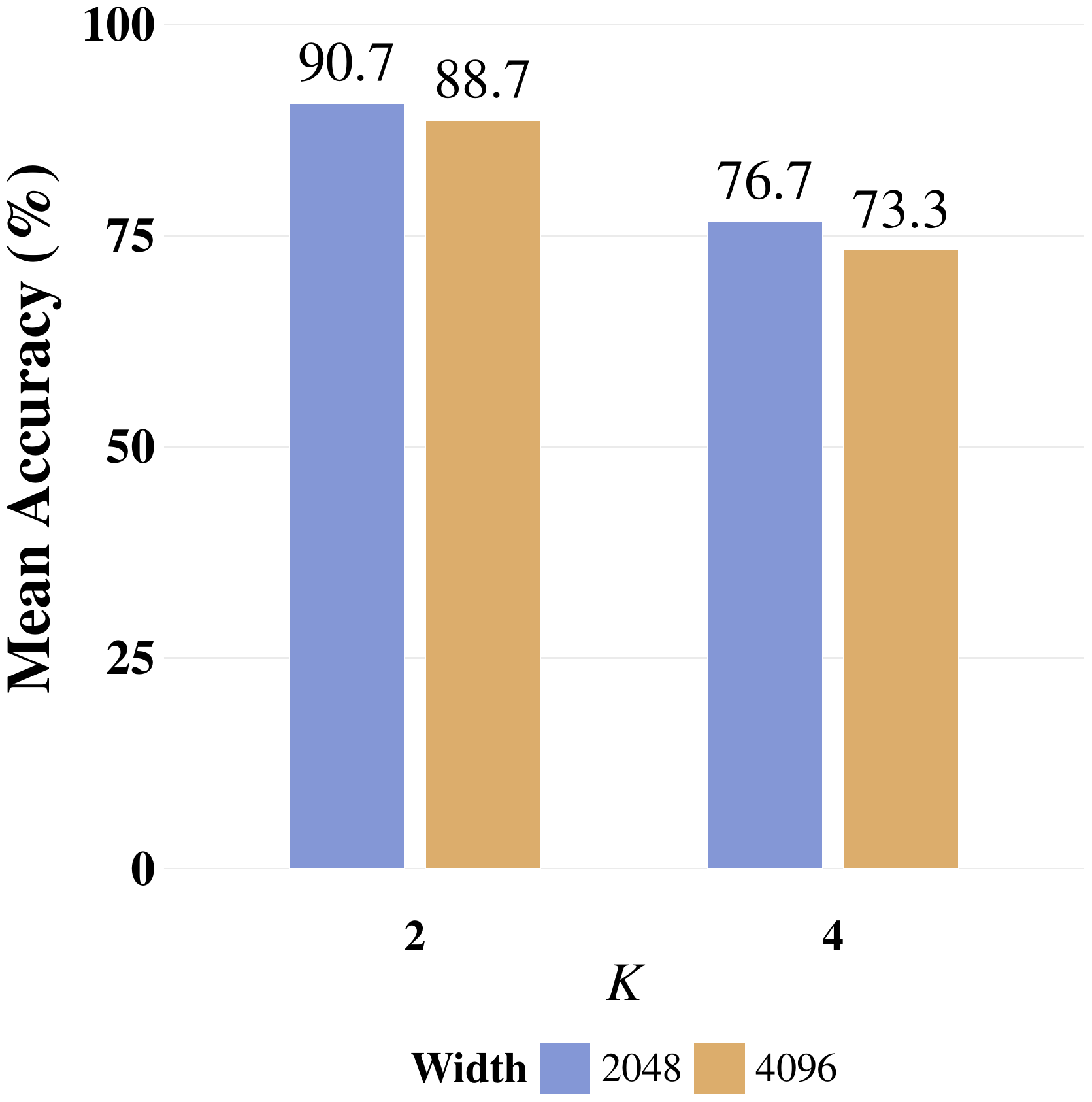}
    \caption{Analysis of graph neighborhood size $K$ and MLP model size with width.}
    \label{fig:k_width}
\end{wrapfigure}
We further analyze the effects of graph connectivity and model capacity on PushT by performing an ablation over $K \in\{2,4\}$ and the MLP model size $\in\{2048,4096\}$, with the results shown in Figure~\ref{fig:k_width}. Each experiment is evaluated with three seeds and 50 episodes per seed. The results show that graph $K$ is the dominant factor. Increasing $K$ from 2 to 4 consistently reduces the mean success rate by approximately 15 percentage points, from 90.7 to 76.7 with the smaller MLP configuration and from 88.7 to 73.3 with the larger MLP configuration. In contrast, increasing the MLP capacity by enlarging its hidden dimension from 2048 to 4096 results in only a modest but consistent performance drop under both $K$ settings, indicating that model capacity is already sufficient at width 2048 and further expansion provides no benefit.
\begin{figure}
  \centering
  \begin{subfigure}[b]{0.24\textwidth}
    \centering
    \includegraphics[width=\linewidth]{figs/heatmap.pdf}
  \end{subfigure}
  \hfill
  \begin{subfigure}[b]{0.24\textwidth}
    \centering
    \includegraphics[width=\linewidth]{figs/heatmap_counterfactual_v2.pdf}
  \end{subfigure}
  \hfill
  \begin{subfigure}[b]{0.24\textwidth}
    \centering
    \includegraphics[width=\linewidth]{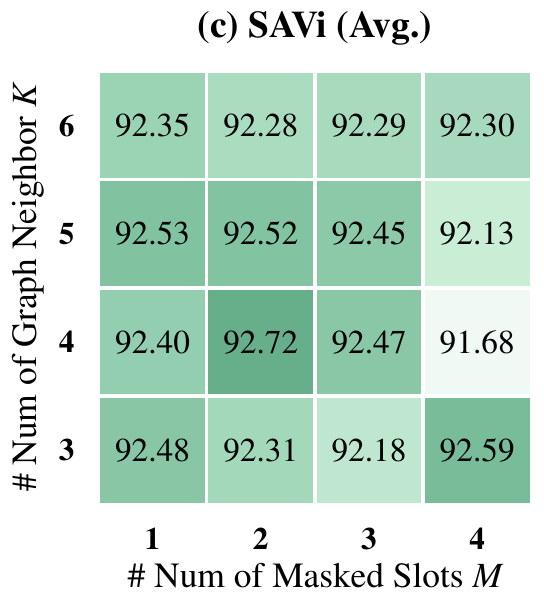}
  \end{subfigure}
  \hfill
  \begin{subfigure}[b]{0.24\textwidth}
    \centering
    \includegraphics[width=\linewidth]{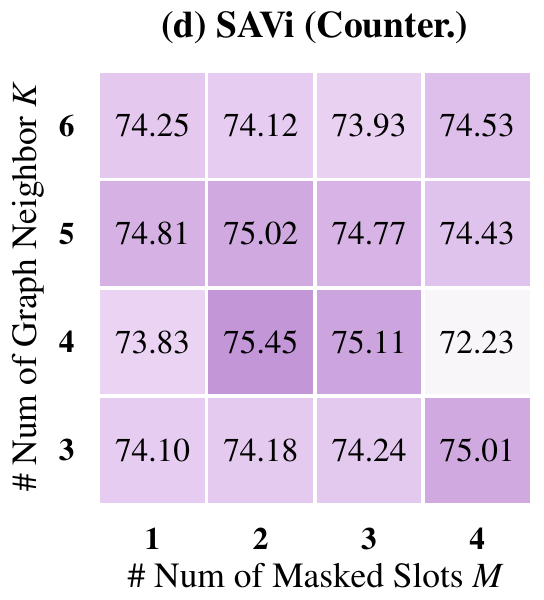}
  \end{subfigure}
  \caption{Analysis of the number of masked objects $M$ and graph neighborhood size $K$. \scriptsize{Average per-question and counterfactual VQA accuracy (\%) are reported under VideoSAUR and SAVi representations.}}
  \label{appx:mk_sensitivity}
\end{figure}
\begin{figure*}[!t]
    \centering

    \begin{subfigure}[t]{0.24\textwidth}
        \centering
        \includegraphics[width=\linewidth]{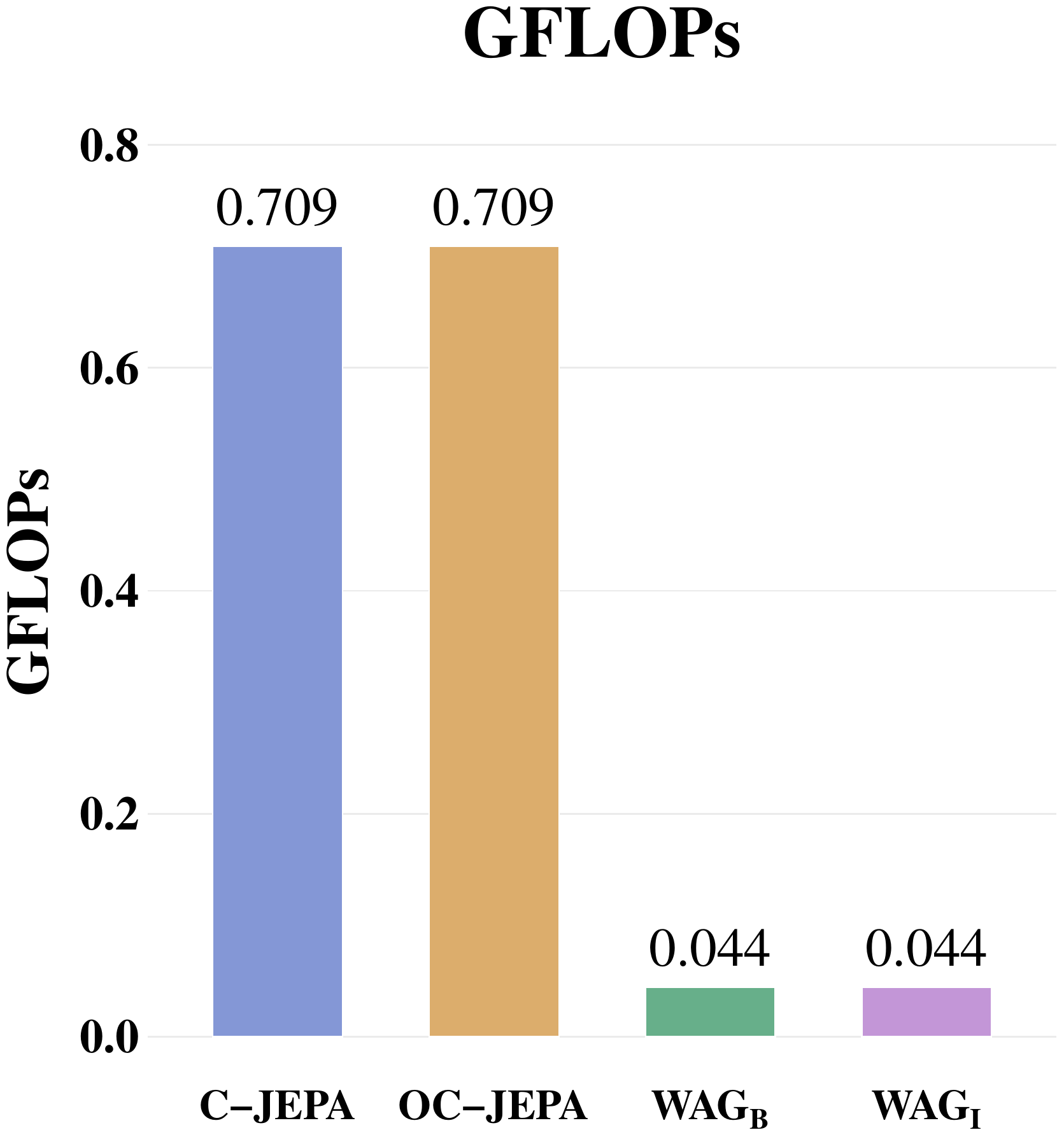}
        \caption{GFLOPs.}
        \label{fig:gflops}
    \end{subfigure}
    \hfill
    \begin{subfigure}[t]{0.24\textwidth}
        \centering
        \includegraphics[width=\linewidth]{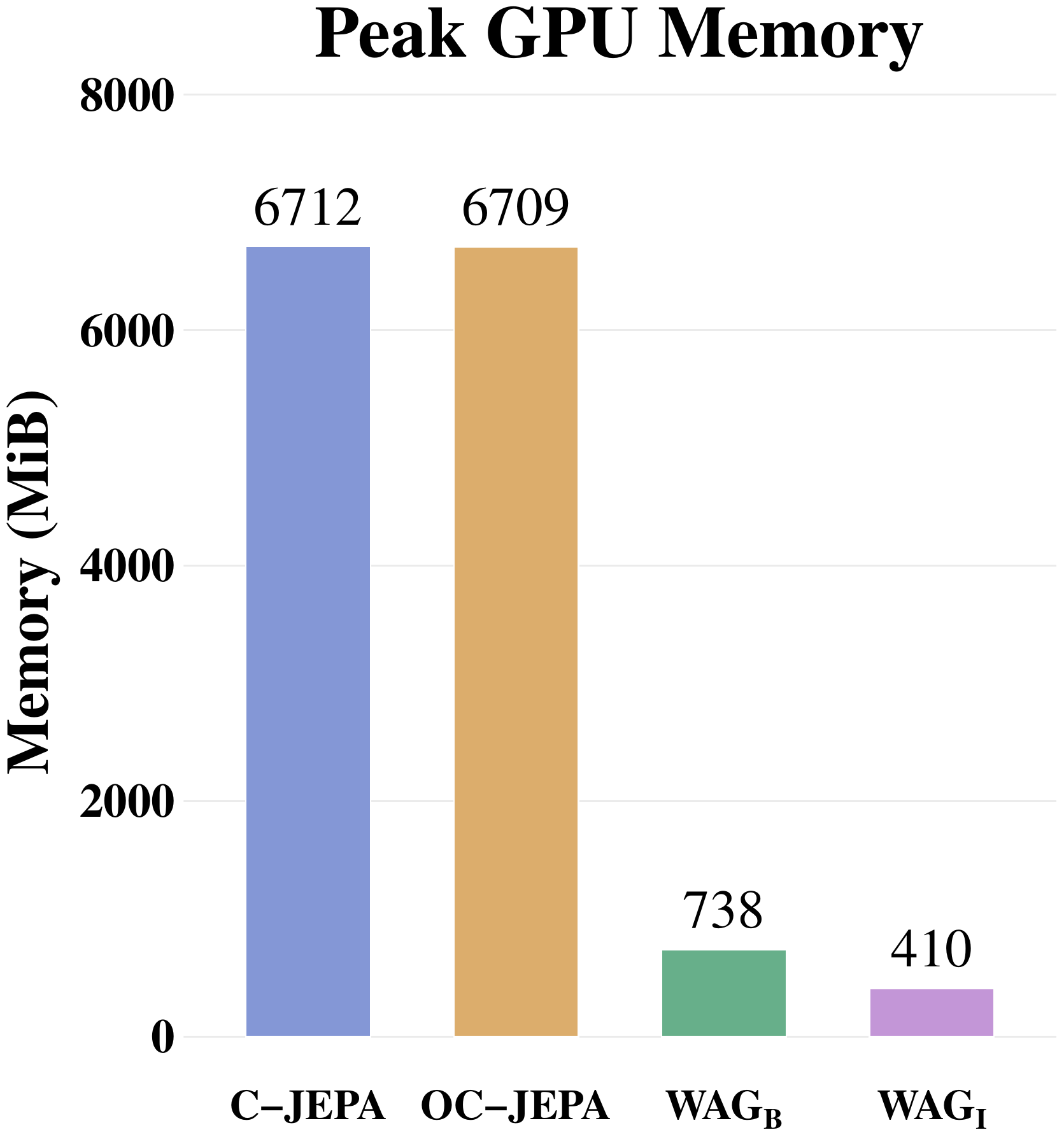}
        \caption{Memory usage.}
        \label{fig:memory}
    \end{subfigure}
    \hfill
    \begin{subfigure}[t]{0.24\textwidth}
        \centering
        \includegraphics[width=\linewidth]{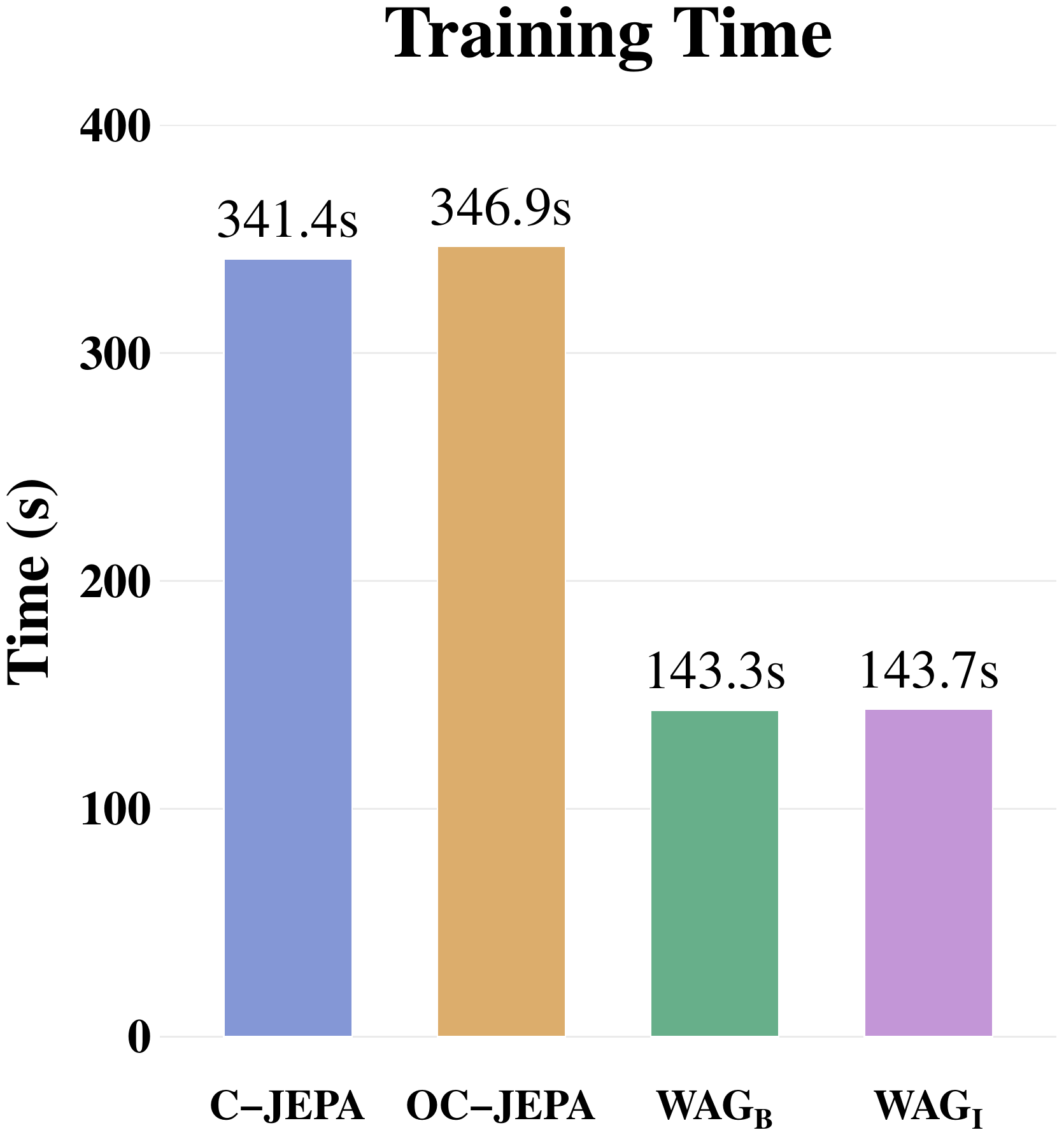}
        \caption{Running time.}
        \label{fig:time}
    \end{subfigure}
    \hfill
    \begin{subfigure}[t]{0.24\textwidth}
        \centering
        \includegraphics[width=\linewidth]{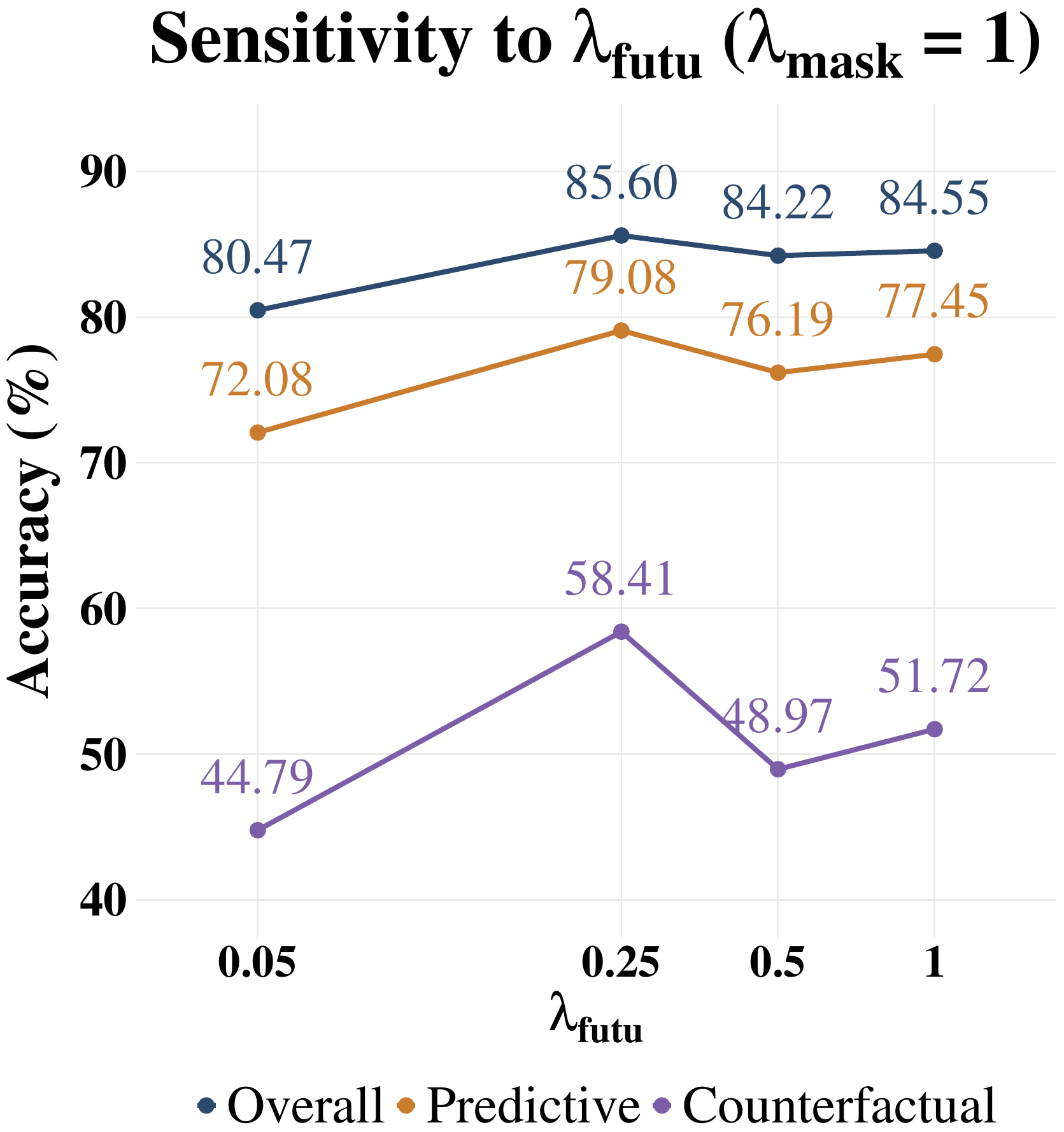}
        \caption{Sensitivity to $\lambda_\text{futu}$.}
        \label{fig:lambda_sensitivity}
    \end{subfigure}

    \caption{Computational efficiency and hyperparameter sensitivity analysis.
    (a) Computational cost measured in GFLOPs.
    (b) Memory consumption.
    (c) Running time.
    (d) Performance sensitivity to the hyperparameter $\lambda_{\mathrm{futu}}$.}
    \label{appx:efficiency_sensitivity}
\end{figure*}

\paragraph{Computational Complexity.} 
\method has a history window of length $T_h$ and prediction horizon $T_p$, with $N$ object slots of dimensionality $D$ and $K < N$ nearest neighbors per node. Thus, the cost of our predictor can be dominated by (a) KNN graph construction via cosine similarity, $O(N^2 D)$; (b) graph message passing over $K$ neighbors, $O(NKD + ND^2)$; and (c) the GRU-based memory update, $O(ND^2)$. Summed over $T$ steps, the total complexity is 
\[
O\big(T(N^2D + NKD + ND^2)\big) = O\big(TND(N+K+D)\big).
\]
Since $D \gg N > K$ in our setting ($D{=}128$, $N{=}7$, $K{\le}6$), this simplifies to $O(TND^2)$, linear complexity in both the number of objects $N$ and the sequence length $T$. This contrasts with a joint object-temporal Transformer predictor such as C-JEPA, which attends jointly over all $TN$ object-timestep tokens and thus incurs $O((TN)^2 D_{\text{model}} + TN\, D_{\text{model}}D_{\text{mlp}})$  quadratic in $TN$. This gap is empirically reflected in a 16$\times$ reduction in FLOPs per sample and 9-16$\times$ lower peak memory for our predictor relative to C-JEPA and OC-JEPA (Figure~\ref{appx:efficiency_sensitivity}).

\paragraph{Comprehensive Results.}
We provide the comprehensive results in the main submissions here, including: (1) Full overall ablation study results in Table~\ref{appx:component_ablation}; (2) Full comparison of different object masking strategies in Table~\ref{appx:mask_strategy}; (3) Full VQA accuracy comparison of different random-graph sampling strategies in Table~\ref{appx:random_graph_comparison}; (4) Full analysis of the number of masked objects $M$ and graph neighborhood size $K$ in Figure~\ref{appx:mk_sensitivity}; (5) Full running costs and hyperparameter analysis in Figure~\ref{appx:efficiency_sensitivity}.

\paragraph{}

\end{document}